\pdfoutput=1

\documentclass[11pt]{article}

\usepackage{acl}

\usepackage{times}
\usepackage{latexsym}

\usepackage[T1]{fontenc}

\usepackage[utf8]{inputenc}

\usepackage{microtype}

\usepackage{amsmath}
\usepackage{graphics}
\usepackage{epsfig}
\usepackage{multirow}
\usepackage{xspace}
\usepackage{subfigure}
\usepackage{caption}
\usepackage{fancyhdr}

\usepackage{subfigure}
\usepackage{multicol}
\usepackage{multirow}
\usepackage{booktabs}
\usepackage{colortbl}
\usepackage{graphicx}

\newcommand{\dataset}{MTG\xspace}

\title{\dataset: A Benchmark Suite for Multilingual Text Generation}

\author{Yiran Chen\textsuperscript{\rm 1}, 
~Zhenqiao Song\textsuperscript{\rm 1}\thanks{\ \  Corresponding author.}, 
~Xianze Wu\textsuperscript{\rm 1}, 
~Danqing Wang\textsuperscript{\rm 1}, \\ 
\textbf{Jingjing Xu\textsuperscript{\rm 1}, Jiaze Chen\textsuperscript{\rm 1}, Hao Zhou\textsuperscript{\rm 2}\thanks{~~Work is done while at ByteDance.}, Lei Li\textsuperscript{\rm 3}\footnotemark[2]} \\
        \textsuperscript{\rm1} ByteDance AI Lab ~~~~~~~~~~  \textsuperscript{\rm2} Insititute for AI Industry Research, Tsinghua University\\
        \textsuperscript{\rm3} University of California, Santa Barbara\\ 
        \texttt{\{chenyiran.robin,songzhenqiao,wuxianze.0\}@bytedance.com} \\
        \texttt{\{wangdanqing.122,chenjiaze\}@bytedance.com} \\
        \texttt{jingjingxu@pku.edu.cn} ~~~~~ \texttt{zhouhao@air.tsinghua.edu.cn}\\
        \texttt{leili@cs.ucsb.edu}\\}

\begin{document}
\maketitle
\begin{abstract}

We introduce \dataset, a new benchmark suite for training and evaluating multilingual text generation. It is the first-proposed multilingual multiway text generation dataset with the largest human-annotated data (400k).
It includes four generation tasks (story generation, question generation, title generation and text summarization) across five languages (English, German, French, Spanish and Chinese). 
The multiway setup enables testing knowledge transfer capabilities for a model across languages and tasks.
Using \dataset, we train and analyze 
several popular multilingual generation models from different aspects. 
Our benchmark suite fosters model performance enhancement with more human-annotated parallel data. It provides comprehensive evaluations with diverse generation scenarios. Code and data are available at \url{https://github.com/zide05/MTG}.
\end{abstract}

\section{Introduction}


Natural language generation (NLG) aims to automatically generate meaningful texts with the input in different formats, such as images~\cite{anderson2018bottom}, tables~\cite{ye2020variational} or texts~\cite{guan2019story}.
The generated texts generally target at realizing an underlying communicative goal while remaining coherent with the input information and keeping grammatically correct.
Multilingual text generation extends the natural language generation task to produce texts in multiple languages, which is important to overcome language barriers and enable universal information access for the world's citizens
 ~\cite{artetxe2019cross,arivazhagan2019massively,pan2021contrastive}.

To achieve this goal, various multilingual text generation datasets have been proposed. 
Some of them do not incorporate cross-lingual pairs~\cite{liang2020xglue,ladhak2020wikilingua}.
This limits the knowledge transfer from one language to another.
Others involve cross-lingual pairs while English is included on either source or target side in most cases~\cite{zhu2019ncls,ladhak2020wikilingua}, leading to difficult transfer between low-resource or distant language pairs.
Constructing a multilingual text generation dataset that can directly transfer knowledge between any two languages is still under-explored.

To this end, we propose \dataset, a human-annotated multilingual multiway dataset. Multiway means that the same sample is expressed in multiple languages.
It covers four generation tasks (story generation, question generation, title generation and text summarization) across five languages (English, German, French, Spanish and Chinese). 
We do not include multilingual machine translation because MT itself is a standard task.
The multiway parallel feature enables cross-lingual data construction between arbitrary language pairs.
Such direct parallel signal promotes knowledge transfer and cross-lingual generation between any language pairs (even distant pairs such as Spanish-Chinese) without involving an intermediate language such as English~\cite{leng2019unsupervised}.

The multilingual multiway feature also enables various training and test scenarios.
In this paper, we design four scenarios to verify the advantages of our MTG from different aspects.
Several representative pretrained multilingual models are employed to test these scenarios, including multilingual BERT (M-BERT)~\cite{devlin2019bert}, XLM~\cite{lample2019cross}, mBART~\cite{liu2020multilingual} and mT5~\cite{xue2020mt5}.
We leverage various metrics to assess the coherence and diversity of the outputs generated by these models.
Besides, we also propose an ensemble metric, which mainly focuses on relevance, measuring to what degree is the generated text close to human-level.
Human evaluation is also conducted to validate models' performances.


In summary, the contributions of this paper are listed as follows: 

\noindent(i) We propose a new human-annotated multilingual multiway text generation benchmark suite \dataset.

\noindent(ii) We design a new evaluation metric measuring how a text resembles human writing 
and prove that it has higher correlation scores with human scores compared with other automatic relevance metrics.

\noindent(iii) We evaluate several representative pretrained multilingual models on our proposed \dataset and make a rigorous analysis to verify its advantages.

\section{Related Work}

A significant body of works have been committed to the construction of multilingual datasets covering diverse tasks~\cite{hu2020xtreme,jiang2020x,longpre2020mkqa}.
XTREME~\cite{hu2020xtreme} is a multilingual understanding benchmark across $40$ languages and $9$ tasks, but it does not cover any generation task.
\citet{jiang2020x} propose X-FACTR, which is a cross-lingual factual retrieval benchmark.
\citet{longpre2020mkqa} propose MKQA, an open-domain question answering evaluation dataset covering $26$ diverse languages.
\citet{ladhak2020wikilingua} present WikiLingua, which is a large-scale, multilingual dataset for cross-lingual abstractive summarization systems. MLSUM~\cite{wang2021contrastive} is a dataset for text summarization in 12 languages. 
Wiki-$40$B~\cite{guo2020wiki} is a multilingual language model dataset across $40+$ languages.
Although these datasets cover multiple languages, they either belong to natural language understanding tasks or a single, specific generation task, which limits researchers to obtain general findings incorporating a set of generation tasks.

XGLUE~\cite{liang2020xglue} is a cross-lingual benchmark dataset for nine understanding tasks and two generation tasks.
GEM~\cite{gehrmann2021gem} is a newly-presented vision-language dataset covering $11$ image-language and video-language tasks and $32$ languages. These two datasets encompass multiple tasks and languages. However, a remarkable 
difference of our \dataset from XGLUE and GEM is that \dataset focuses on text-to-text generation tasks and is parallel across all languages, which facilitates easier knowledge transfer.


\section{Dataset Collection and Methodology}

This section will introduce how to create the benchmark suite for multilingual text generation (\dataset). 
In order to construct multiway parallel dataset, the initial dataset is translated into other languages by an off-the-shelf translation model. Part of the translated data is randomly selected for further human annotation to increase data quality.
The selection of tasks, initial datasets and languages are based on several principles as described below.
\renewcommand\arraystretch{1}
\begin{table*}
\footnotesize
  \centering
  \resizebox{0.99\textwidth}{!}{
    \begin{tabular}{lllcl}
    \toprule
    \multicolumn{1}{c}{\textbf{Task}} & \multicolumn{1}{c}{\textbf{Corpus}} & \multicolumn{1}{c}{\textbf{Domain}} & \multicolumn{1}{c}{\textbf{Format}} & \multicolumn{1}{c}{\textbf{Goal}} \\
    \midrule
    Story Generation & ROCStories & Daily life & <story> & Generate the end of the story \\
    Question Generation & SQUAD 1.0 &  Wikipedia  & <passage,answer, question> & Generate the question of the answer \\
    Title Generation & ByteCup &  News  & <article, title> & Generate the title of the document \\
    Text Summarization & CNN/DailyMail &  News  & <article, summary> & Generate the summary of the document \\
    \bottomrule
    \end{tabular}%
    }
  \caption{The description of tasks and English datasets included in \dataset. For story generation, we use the last sentence as story end to be generated and the rest as input.}
  \label{tab:tasks}%
\end{table*}%

\subsection{Task and Dataset Selection}
It is important to select suitable tasks for our \dataset benchmark to make it diverse and challenging.
Thus, we define several criteria during the task selection procedure: 

\noindent \textbf{Task Definition} Tasks should be well-defined, which means that humans can easily determine whether the generated results meet the task requirements. 


\noindent \textbf{Task Difficulty} Tasks should be solvable by most college-educated speakers. In the meantime, they should be challenging to current models, the performance of which in various test scenarios falls short of human performance. 

\noindent \textbf{Task Diversity} Tasks should cover a wide range of generation challenges that allow for findings to be as general as possible.

\noindent \textbf{Input Format} The input format of the tasks needs to be as simple as possible to reduce the difficulty of data processing. Besides, it should not contain anything but text (e.g., without any images or videos).




In order to meet the above criteria, 8 domain experts are asked to vote from 10 typical generation tasks\footnote{These generation tasks are story generation, commonsense generation, style transfer, question generation, question answering, dialogue generation, title generation, text summarization, image caption, and data-to-text generation.}. Finally, four generation tasks are selected for \dataset, which are story generation, question generation, title generation and text summarization.
\textbf{Story generation} (SG) aims to generate the end of a given story context, which requires the model to understand the story context and generate a reasonable and fluent ending~\cite{guan2019story}.
\textbf{Question generation} (QG) targets at generating a correct question for a given passage and its answer~\cite{duan2017question}. For the same passage with different answers, the system should be able to generate different questions.
\textbf{Title generation} (TG) converts a given article into a condensed sentence while preserving its main idea~\cite{jin2002new}. The title should be faithful to the original document and encourage users to read the news at the same time.
\textbf{Text summarization} (Summ) aims to condense the source document into a coherent, concise, and fluent summary~\cite{mani2001automatic}. It is similar to title generation but the output of text summarization is relatively longer.
These four tasks focus on different generative abilities and realize different goals.

After confirming the tasks, the next step is to choose the dataset for each task.
The two selection principles are listed as follows:(1) \textbf{License:} Task data must be available under licenses that allow using and redistributing for research purposes. The dataset should be free and available for download. (2) \textbf{Quality:} The dataset size should be as large as possible and the quality should be checked. 

English datasets are chosen as the initial datasets because they are more accessible in all four tasks and have relatively larger size compared with datasets in other languages.
We choose ROCStories~\cite{mostafazadeh2016corpus} for story generation, SQUAD $1.0$~\cite{rajpurkar2016squad} for question generation, ByteCup
\footnote{https://www.biendata.xyz/competition/bytecup2018/} 
for title generation and CNN/DailyMail~\cite{nallapati2016abstractive} for text summarization. These datasets are popular in the corresponding fields and have been verified to be high-quality by many works. Moreover, they are all under a permissive license.
An overview of all task datasets is shown in Table \ref{tab:tasks}.


\subsection{Language Selection}
The original datasets are in \textbf{English} (en) only and we want to extend them into a multiway parallel form. This means that all English texts should be translated into other languages, which will lead to high annotation costs.
Thus, a state-of-the-art translator is leveraged to do the translation and then annotators are asked to correct the translated text. Considering this construction method, \dataset should contain languages that (1) have good English-to-X translators and (2) are diverse in language family. 
Finally, \textbf{German} (de), \textbf{French} (fr), \textbf{Spanish} (es) and \textbf{Chinese} (zh) are chosen. German is from the same language branch as English while French and Spanish are from different ones. Chinese is more distant from the rest of languages in the language family tree.
\renewcommand\arraystretch{1}
\begin{table}[htbp]
\small
  \centering
  
    \begin{tabular}{lr}
    \toprule
    \textbf{Task} & SG, QG, TG, Summ \\
    \midrule
    \multicolumn{2}{l}{\textbf{For each language}} \\
    \midrule
    Rough training size  & 76k/61k/270k/164k \\
    Annotated training size & 15k/15k/15k/15k \\
    Annotated development size & 2k/2k/2k/2k \\
    Annotated test size & 3k/3k/3k/3k \\
    \midrule
    \multicolumn{2}{l}{\textbf{For five languages (en, de, fr, es, zh)}} \\
    \midrule
    Total Annotated size & 400k \\
    Total dataset size & 6.9m \\
    \bottomrule
    \end{tabular}%
  \caption{The number of samples in \dataset. \dataset consists of four subsets: \textit{rough training}, \textit{annotated training}, \textit{development} and \textit{test} set. The rough training set is filtered by back translating across five languages. The annotated training, development and test sets are corrected by human experts.}
  \label{tab:dataset}%
\end{table}%


\begin{table*}[htbp]
  \centering
    \footnotesize
  \resizebox{0.99\textwidth}{!}{
    \begin{tabular}{lccccccccc}
    \toprule
    Correlation & AdaBoost & DecisionTree & ExtraTree & GradientBoosting & Kneighbors & Linear & RandomForest & SVR   & Bagging \\
    \midrule
    Pearson & 0.100  & 0.133  & 0.190  & 0.215  & 0.192  & 0.173  & 0.208  & 0.113  & \textbf{0.240 } \\
    \midrule
    Correlation & BLEU  & ROUGE-1 & ROUGE-2 & ROUGE-L & METEOR & BERTScore-P & BERTScore-R & BERTScore-F1 & Bagging \\
    \midrule
    Pearson & 0.180  & 0.142  & 0.163  & 0.144  & 0.122  & 0.142  & 0.176  & 0.162  & \textbf{0.344 } \\
    \bottomrule
    \end{tabular}%
    }
    \caption{The correlation scores between automatic metric scores and human-annotated scores (the average scores of grammar, fluency and relevance). Upper part of the table shows the correlation scores of different regression algorithms in test set of all languages. The lower part demonstrates correlation scores of our ensemble score (the bagging regressor) and other classic automatic scores in test set without Chinese results because Meteor does not support Chinese. }
  \label{tab:correlation all}%
\end{table*}%

\subsection{Data Collection}
\label{sec: data collection}
After determining the tasks and languages, we introduce the data collection process to get the \dataset.
The Google Translate\footnote{https://translate.google.com/} is used to translate the English datasets to the selected languages. To control the quality of translated texts, we back translate the text to English and filter the samples whose n-gram overlap ratios with the original English texts are lower than a certain threshold.
Different threshold values (from $0.3$ to $0.6$ with $0.1$ as step length) are tested and if it is set to $0.6$, the training data size of QG will drop more than $60\%$. Thus we decide to use $0.5$ as the threshold number to improve the quality of the filtered data while still maintaining more than $70\%$ of the original training data.\footnote{The detailed sizes of the filtered datasets with respect to different thresholds are included in appendix \ref{sec:back translation threshold}.}
Samples in four languages are aligned to ensure that the dataset is multiway parallel. 



$20,000$ samples of each task and language are randomly selected for annotation under the premise of ensuring inter-language alignment.
The annotators are required to further check the translated results based on the following rules:
(1) \textbf{Semantic aligned} Whether the target text is meaningful and is fully semantic aligned with the source text.
(2) \textbf{Fluency} Whether the translated text is grammatically correct.
(3) \textbf{Style} Whether the translation follows the norms of local culture, language conventions, and gender-related words.
If the translated text contradicts any of the above rules, annotators will correct it accordingly. The annotated data is then split to $15$k/$2$k/$3$k as training/development/test subsets. 

A team of 10 full-time experts\footnote{There are 3 language experts for German, 3 for French, 4 for Spanish and 4 for Chinese} are hired 
to do the annotation, who are paid daily. Some part-time workers\footnote{There are 16 part-time workers who are participated in the German annotation, 39 for French, 4 for Spanish and 15 for Chinese.} are also employed 
to increase the annotation throughput, who are paid by the number of annotations. Each annotator is an expert in at least two languages (English and another target language). 
They are trained to correct translation errors according to the above rules, first a small number of samples for trial, these annotation results are re-checked by us and feedback is given to the annotators to help them understand the tasks better.
After this annotation training process, the annotators start to annotate the dataset. 
For quality control, we sample $2\%$ from the annotations and arrange for 9 experts to double-check them. Each example is assigned to two other experts and the data is qualified only if both of them agree on the annotation\footnote{The grammar, expressions, and punctuation of the annotated text are completely correct and the expressions are in accordance with the foreign language.}. If more than $5\%$ of the annotations fail, then all the data of that annotator for that day will be re-checked. 


Then the multiway parallel generation benchmark \dataset is finally completed. It contains four different generation tasks in five languages and its quality is improved by the incorporation of human annotation. However, the number of human-annotated data is still small due to cost concerns. Introducing more human-annotated data or carrying out extra filtering for machine-translated data can be future directions to further improve the quality of \dataset.
The statistics of \dataset is shown in Table \ref{tab:dataset}.

\section{Experiments}

In this section, we conduct extensive experiments to benchmark the difficulty of our proposed \dataset via several state-of-the-art multilingual models under different scenarios.




\begin{figure*}[htbp]
\captionsetup{skip=1pt}
  \centering
  \subfigure[SG M-BERT]{
  \label{fig:SG M-BERT}
  \includegraphics[width=0.17\linewidth]{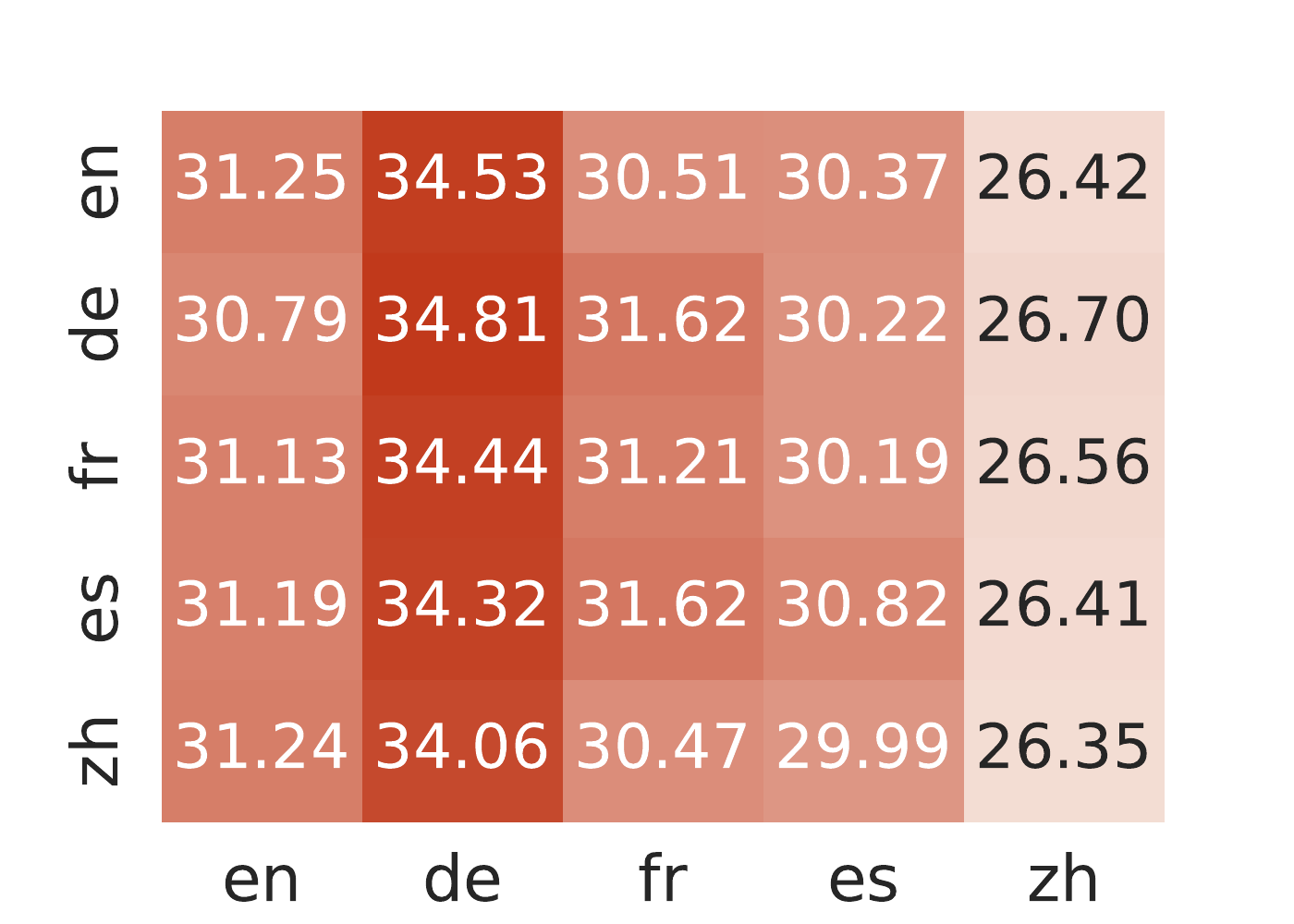}
  }\hspace{-2mm}
  \subfigure[SG XLM]{
  \label{fig:SG XLM}
  \includegraphics[width=0.17\linewidth]{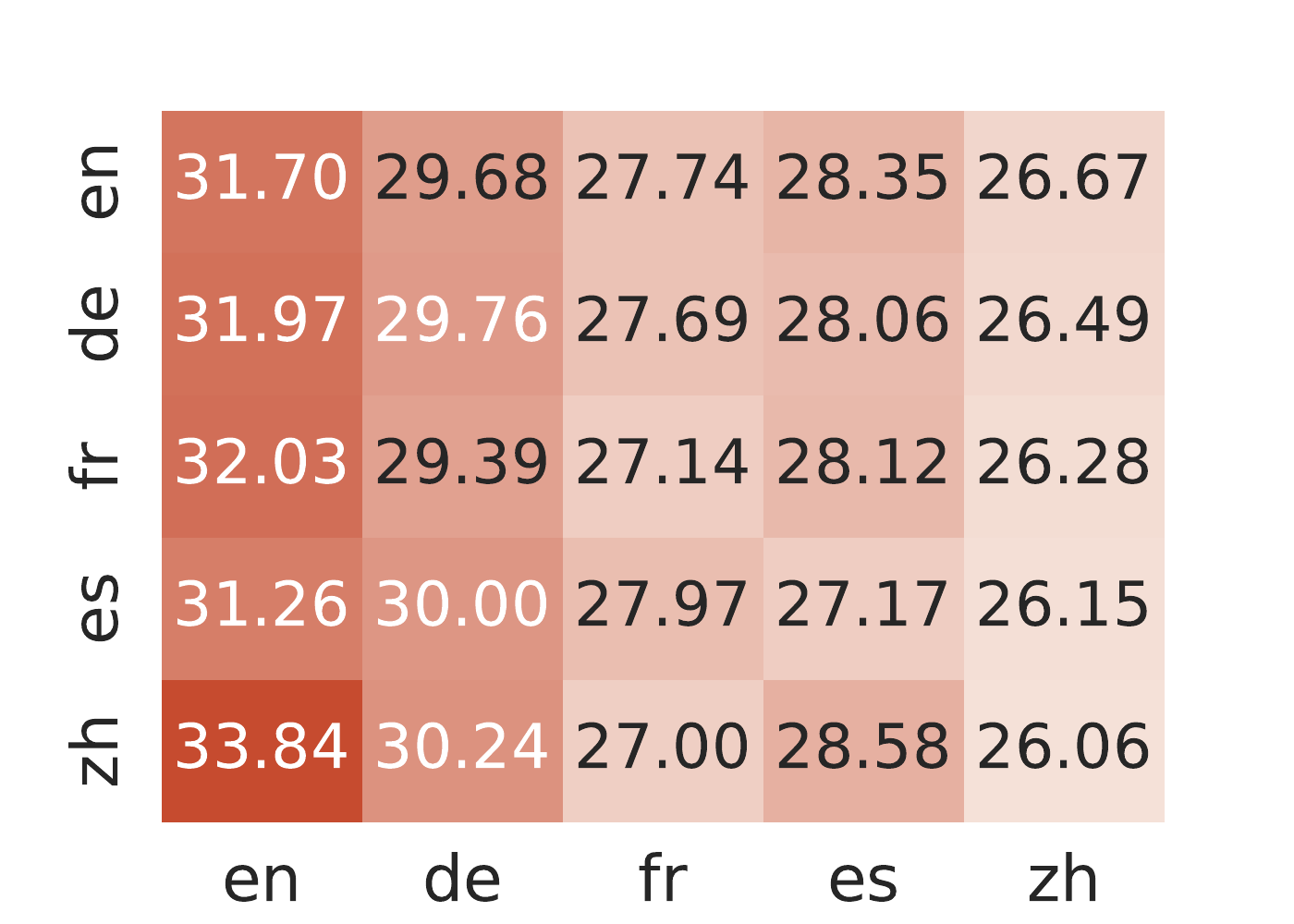}
  }\hspace{-2mm}
  \subfigure[SG mBART]{
  \label{fig:SG mBART}
  \includegraphics[width=0.17\linewidth]{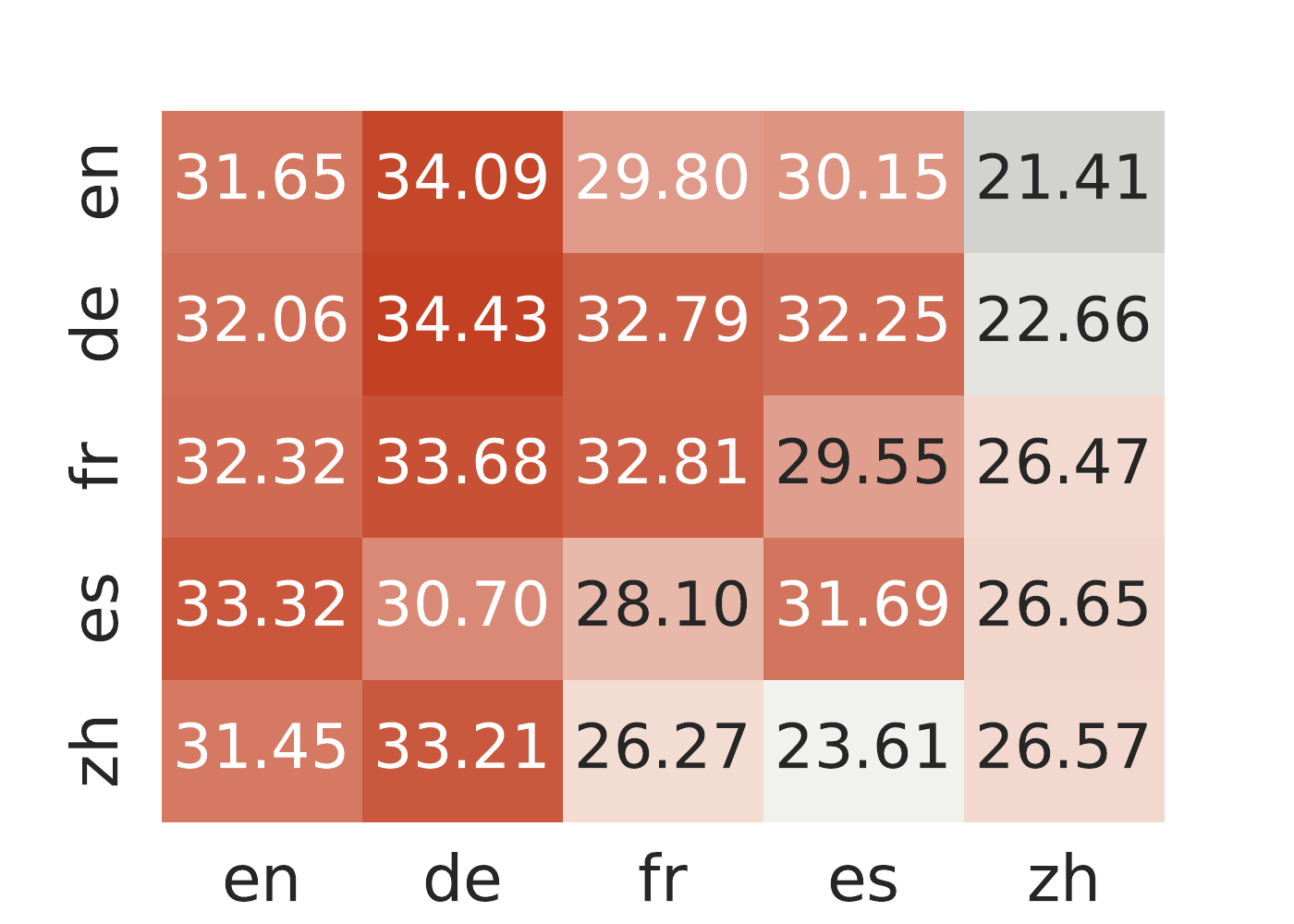}
  }\hspace{-2mm}
  \subfigure[SG mT5]{
  \label{fig:SG mT5}
  \includegraphics[width=0.18\linewidth]{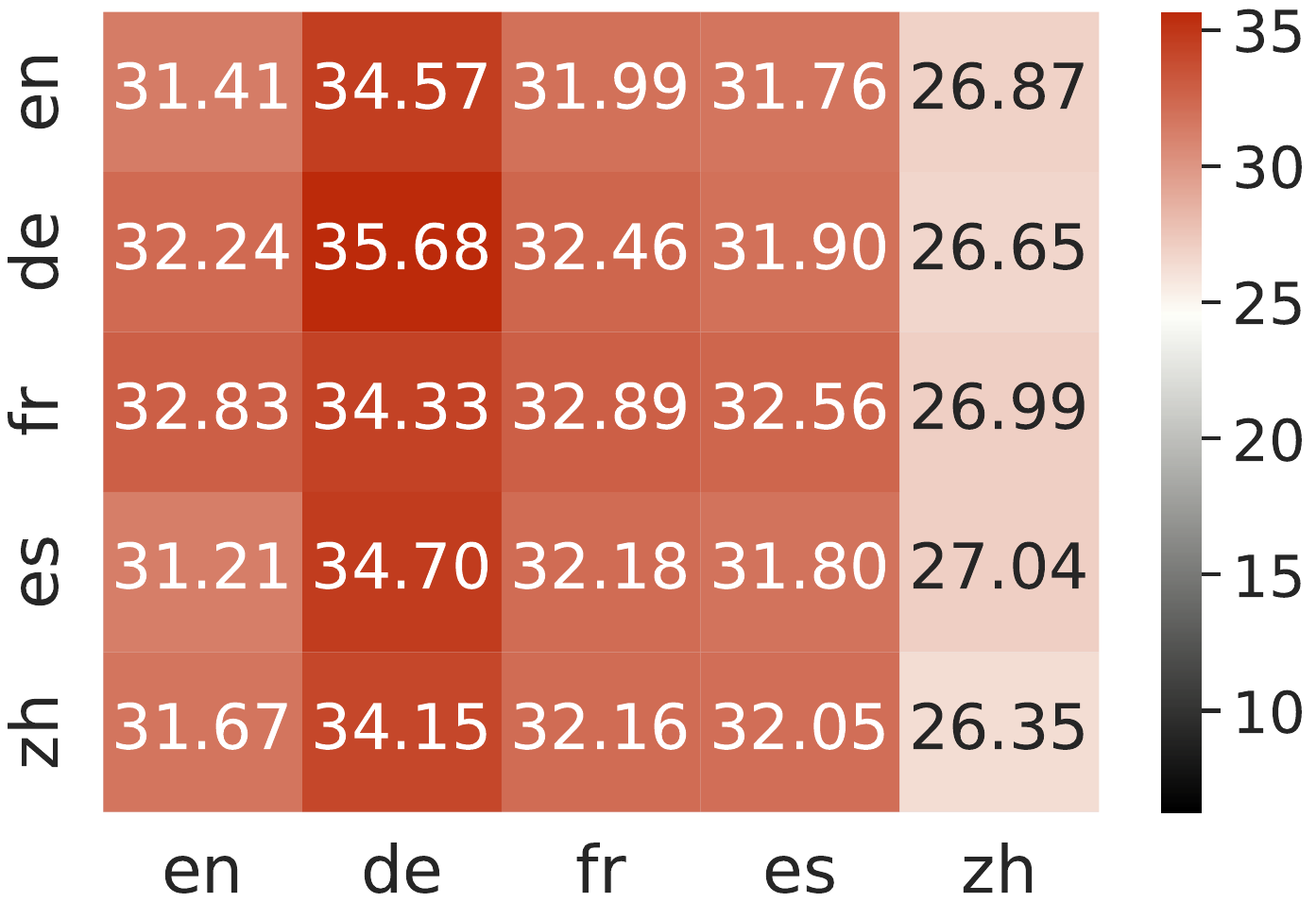}
  }\vspace{-4mm}
  
  \subfigure[QG M-BERT]{
  \label{fig:QG M-BERT}
  \includegraphics[width=0.17\linewidth]{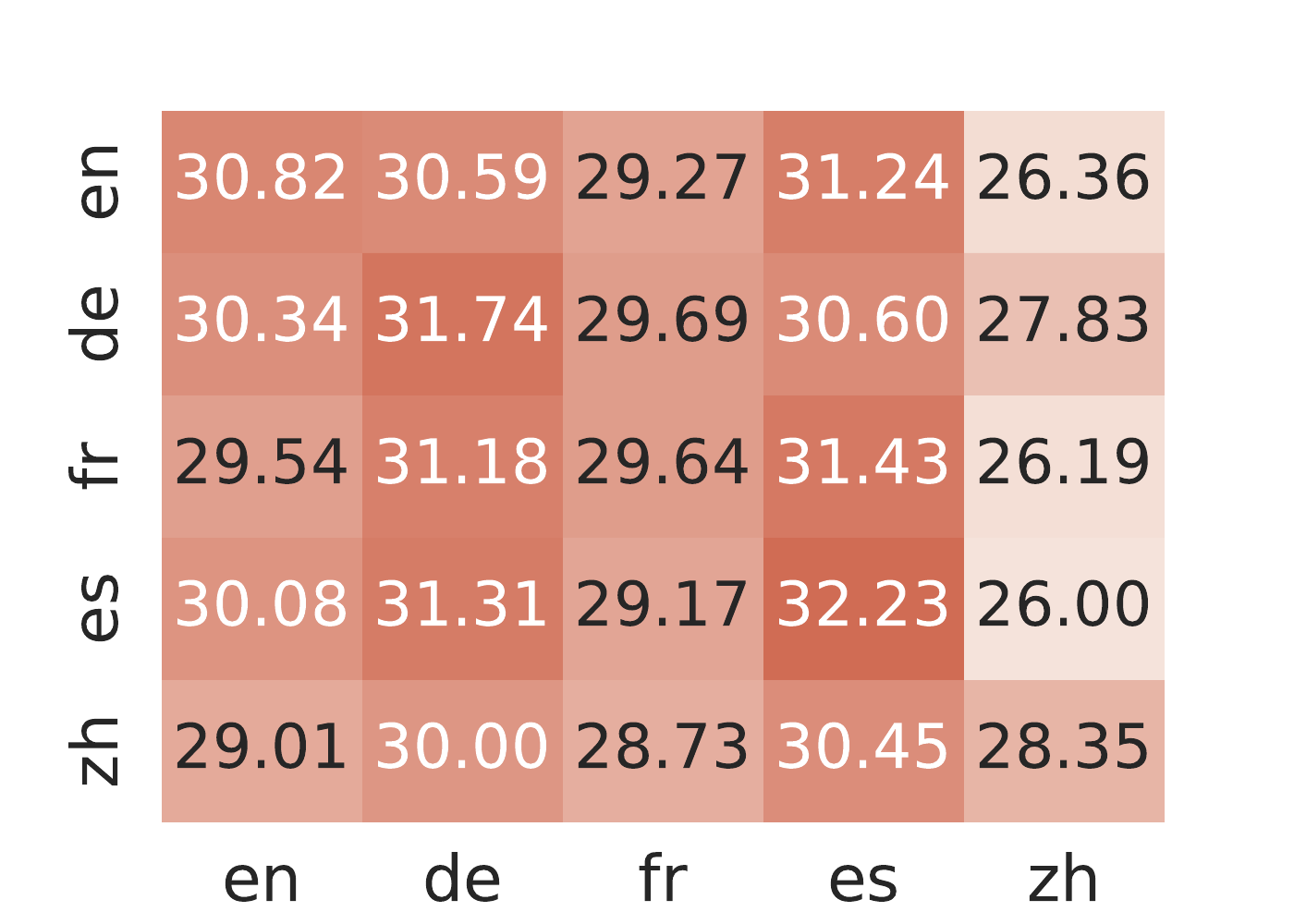}
  }\hspace{-2mm}
  \subfigure[QG XLM]{
  \label{fig:QG XLM}
  \includegraphics[width=0.17\linewidth]{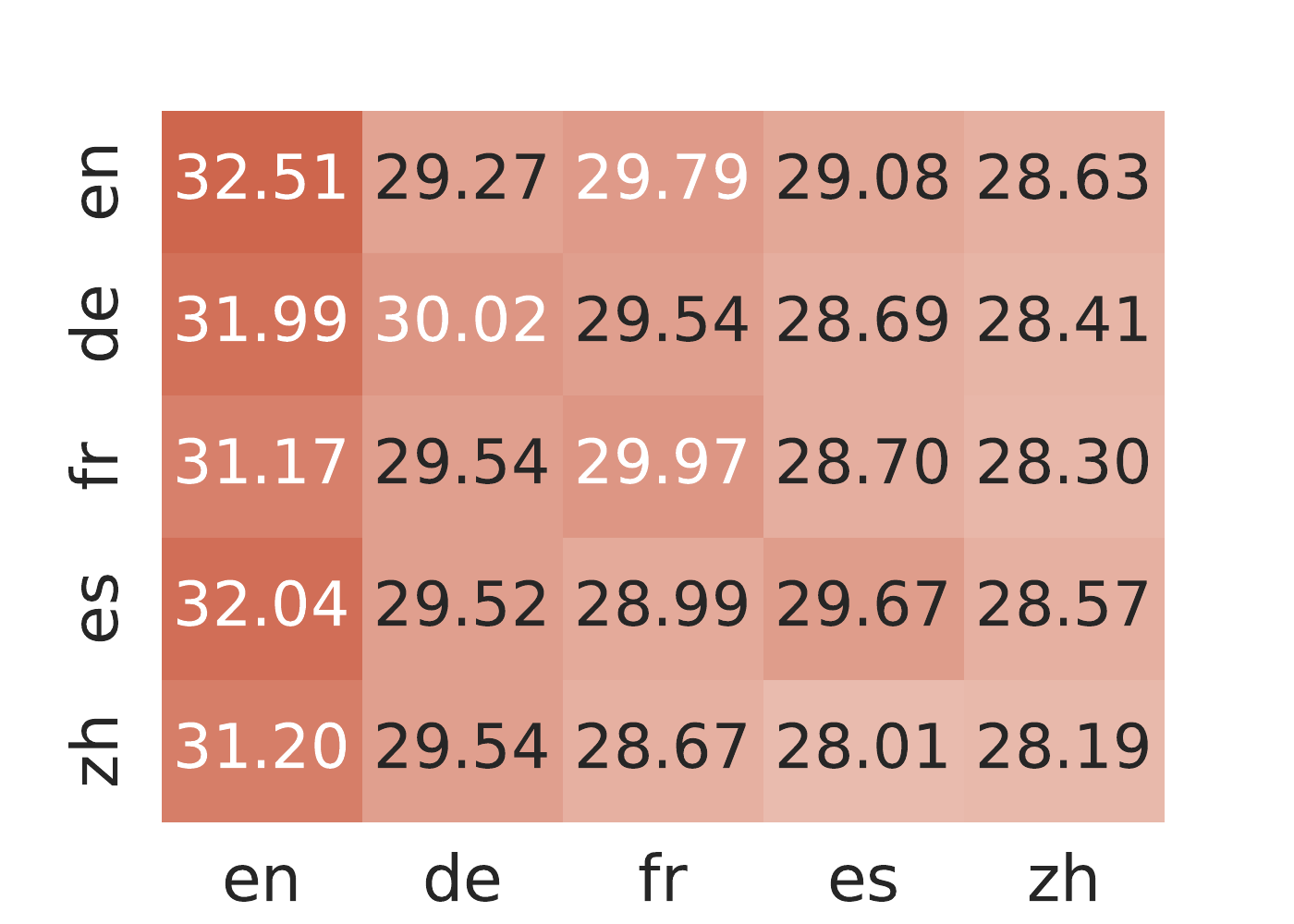}
  }\hspace{-2mm}
  \subfigure[QG mBART]{
  \label{fig:QG mBART}
  \includegraphics[width=0.17\linewidth]{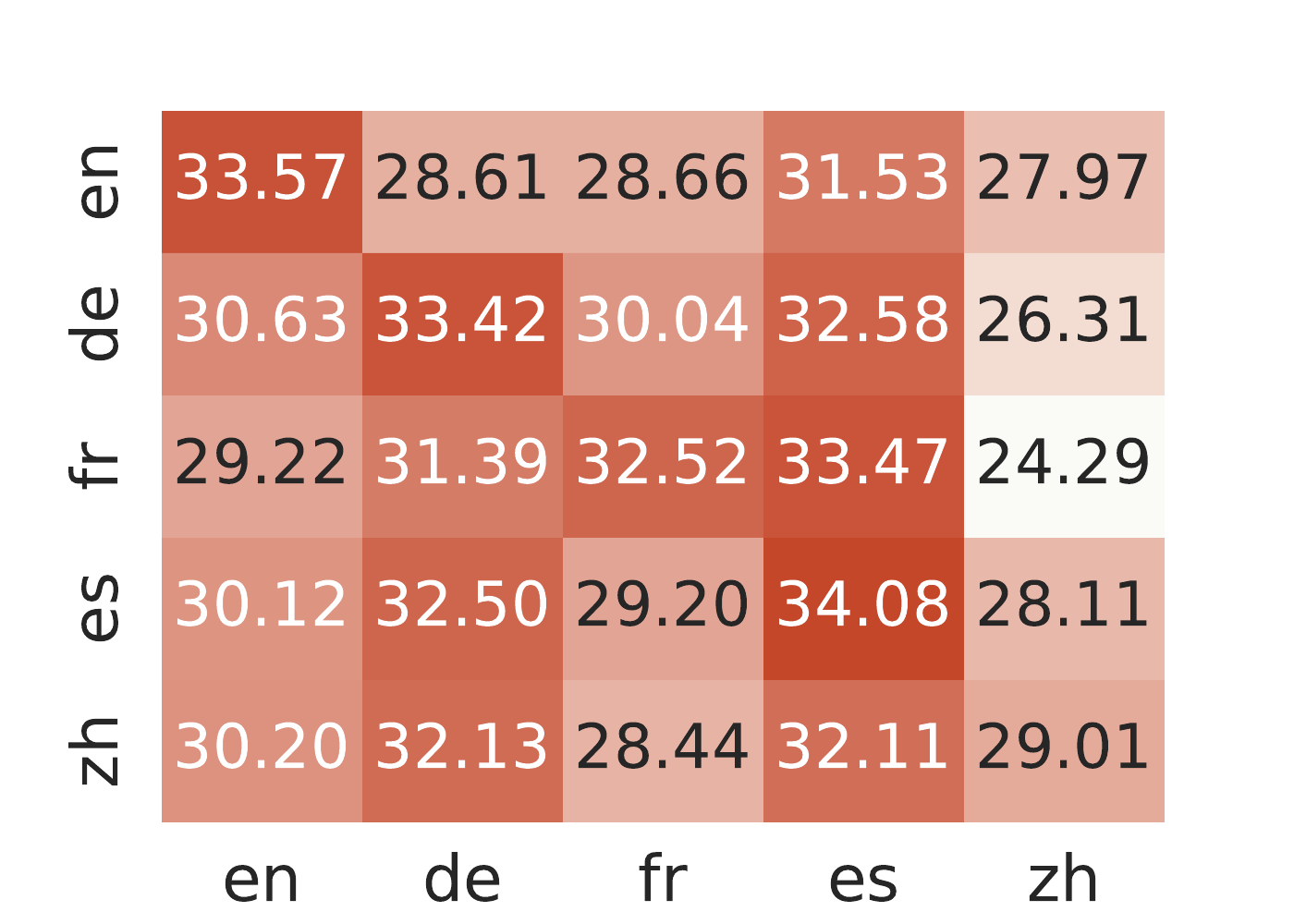}
  }\hspace{-2mm}
  \subfigure[QG mT5]{
  \label{fig:QG mT5}
  \includegraphics[width=0.18\linewidth]{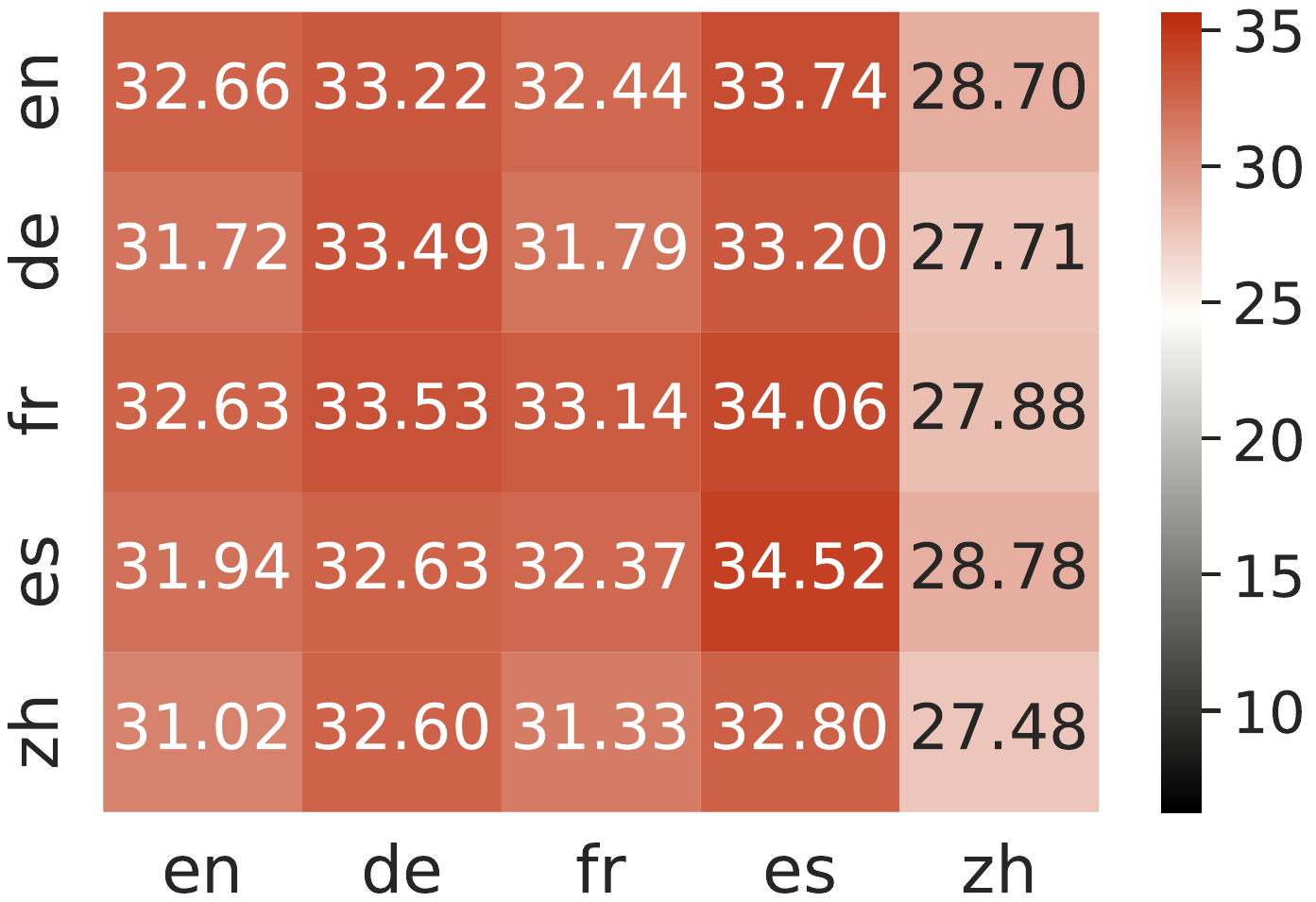}
  }\vspace{-4mm}
  
  \subfigure[TG M-BERT]{
  \label{fig:TG M-BERT}
  \includegraphics[width=0.17\linewidth]{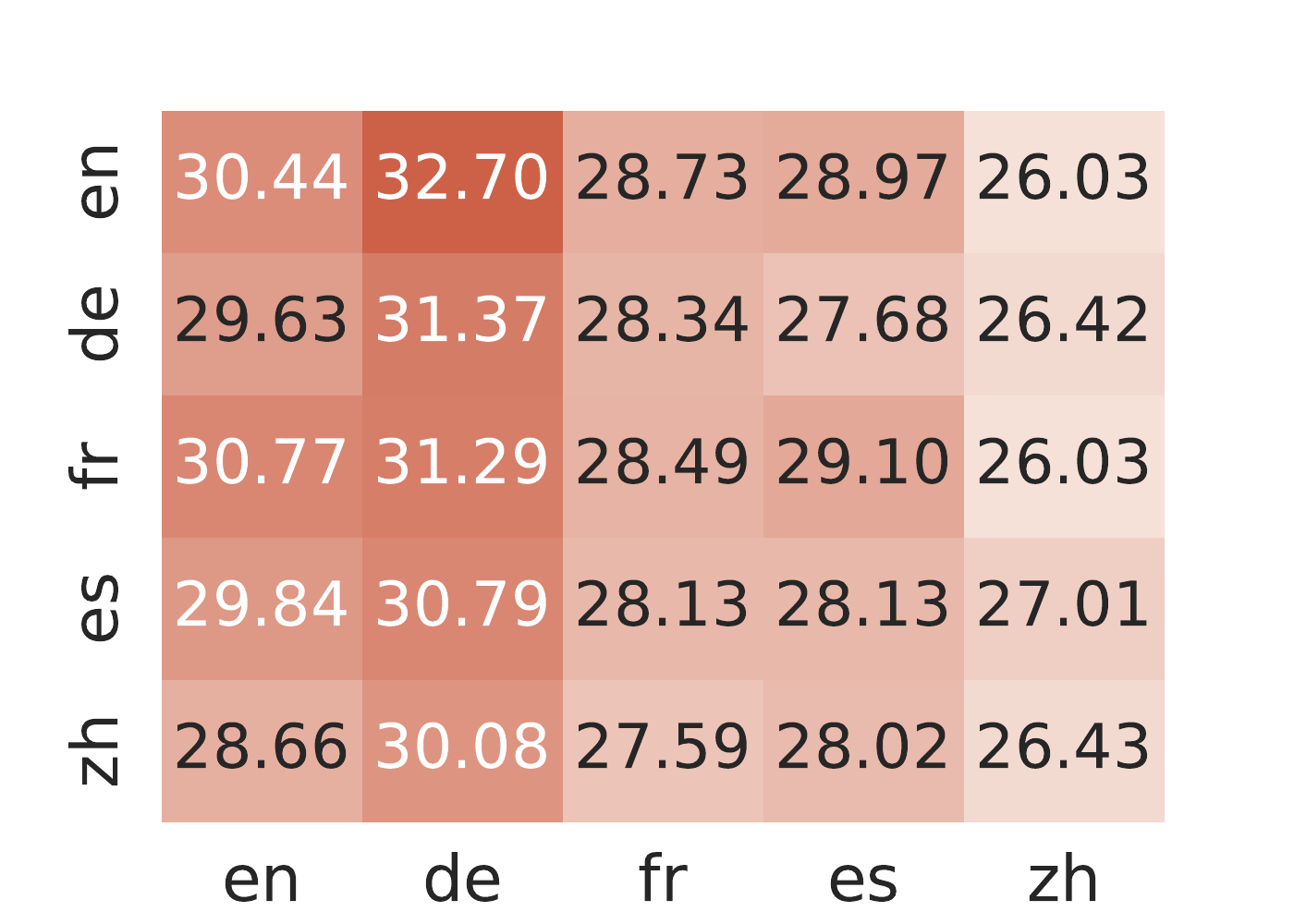}
  }\hspace{-2mm}
  \subfigure[TG XLM]{
  \label{fig:TG XLM}
  \includegraphics[width=0.17\linewidth]{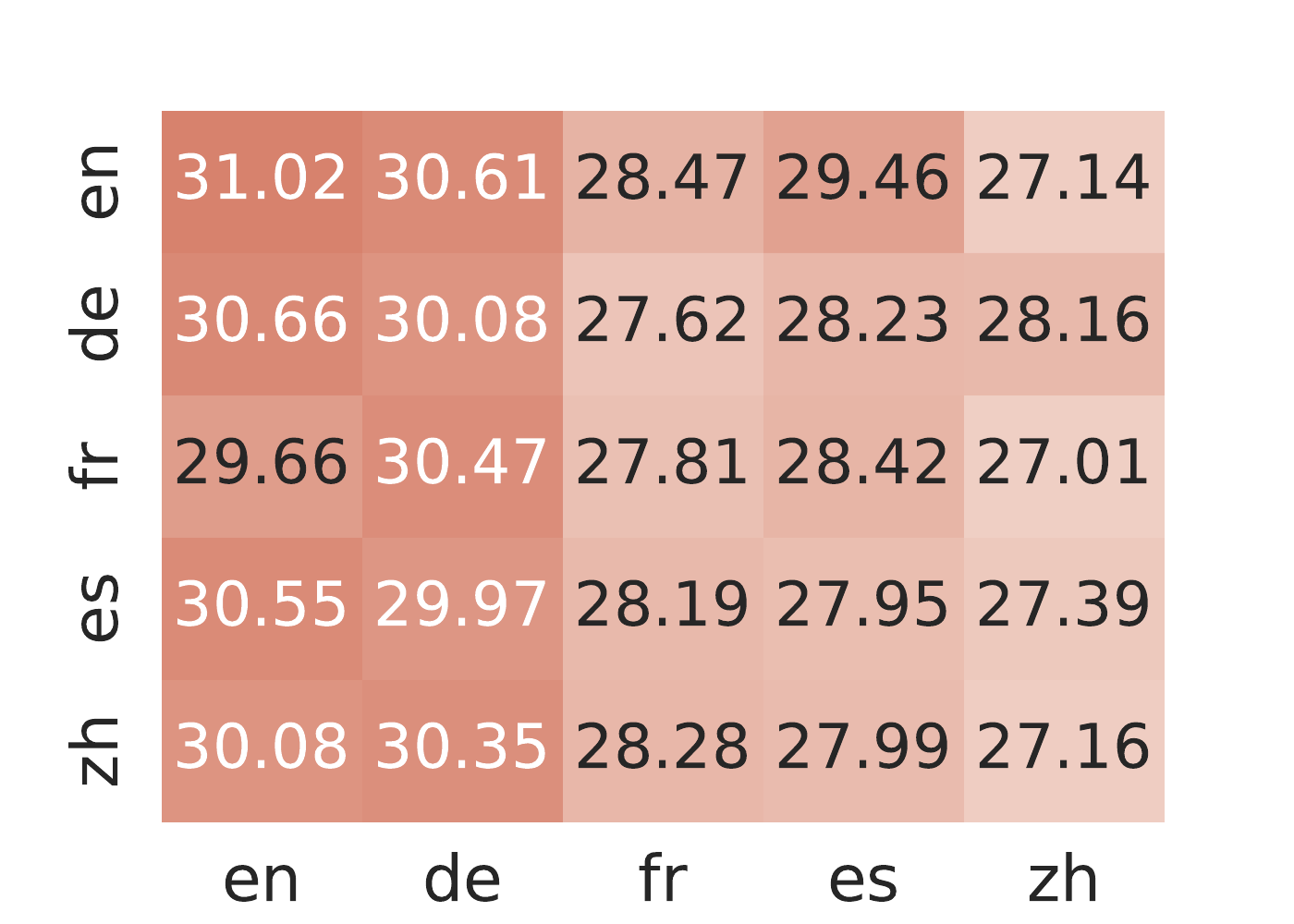}
  }\hspace{-2mm}
  \subfigure[TG mBART]{
  \label{fig:TG mBART}
  \includegraphics[width=0.17\linewidth]{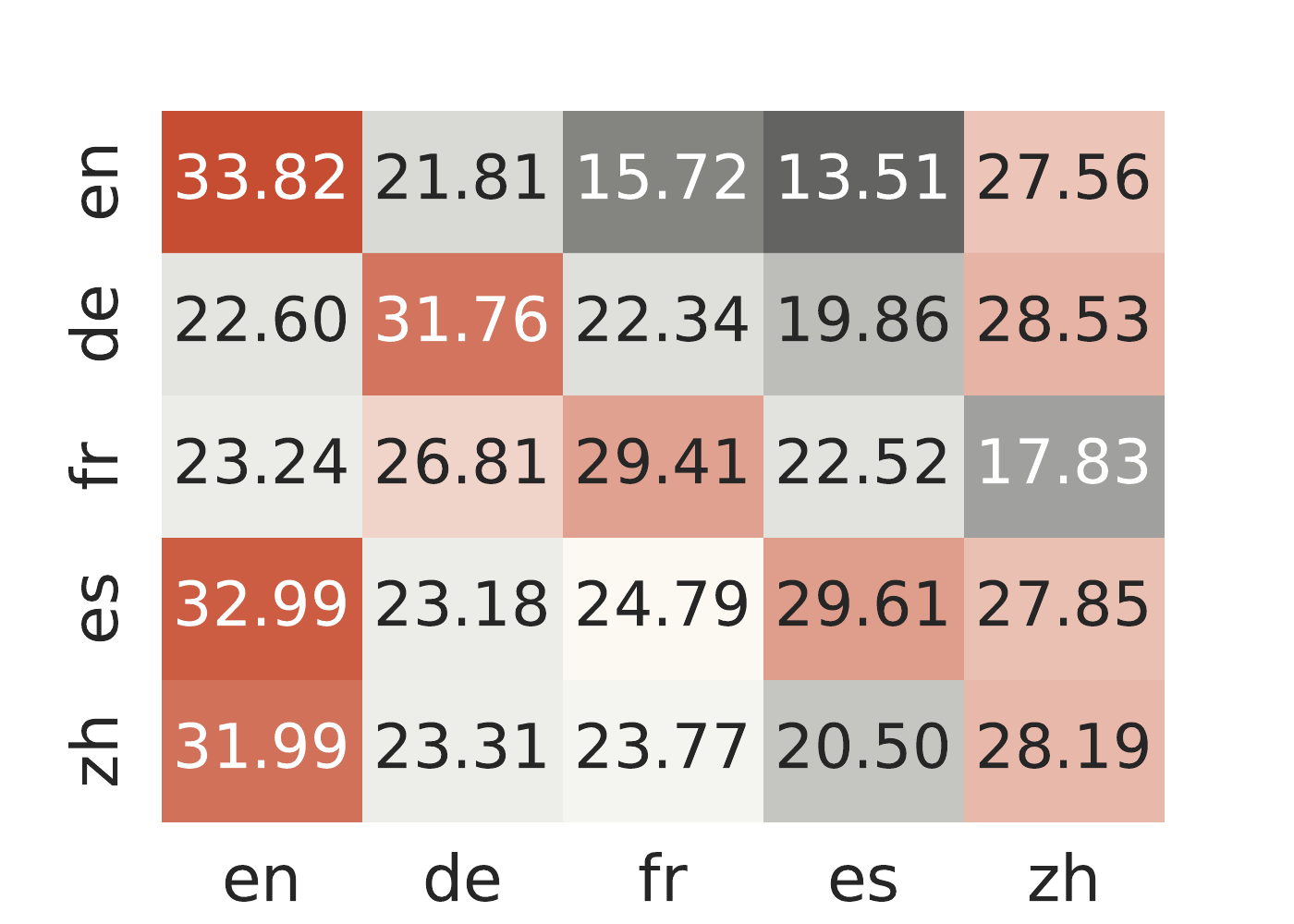}
  }\hspace{-2mm}
  \subfigure[TG mT5]{
  \label{fig:TG mT5}
  \includegraphics[width=0.18\linewidth]{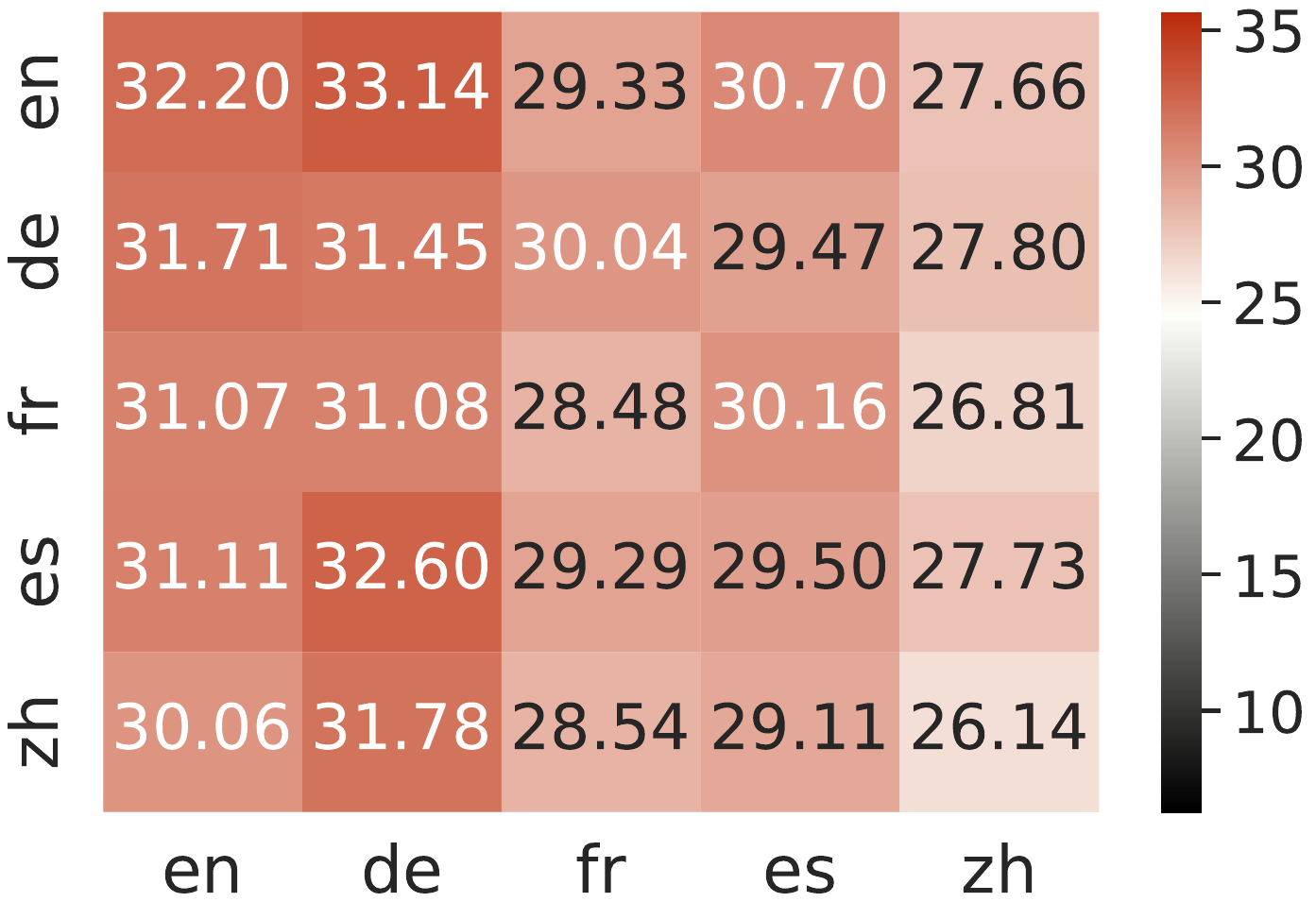}
  }\vspace{-4mm}

  \subfigure[Summ M-BERT]{
  \label{fig:Summ M-BERT}
  \includegraphics[width=0.17\linewidth]{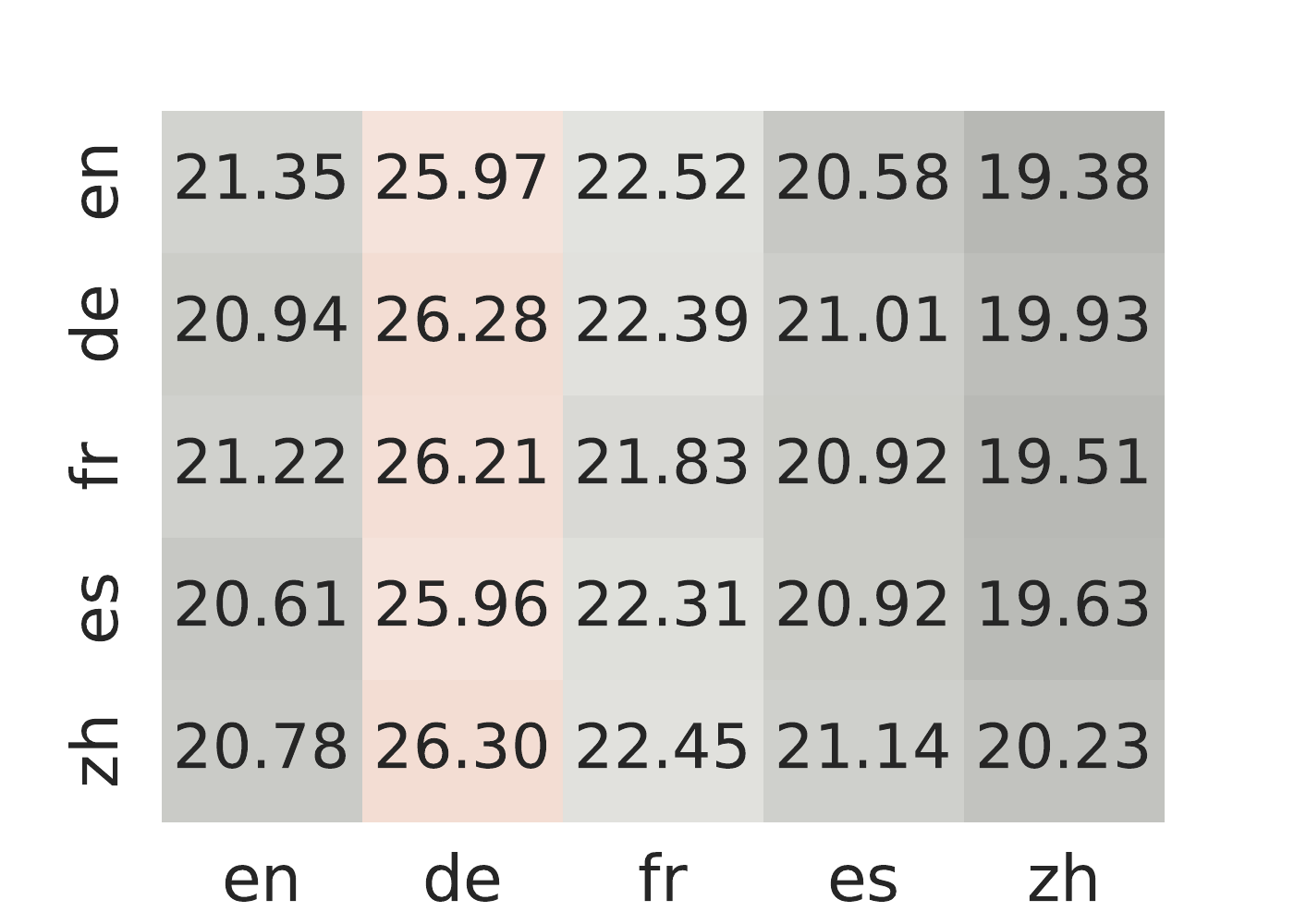}
  }\hspace{-2mm}
  \subfigure[Summ XLM]{
  \label{fig:Summ XLM}
  \includegraphics[width=0.17\linewidth]{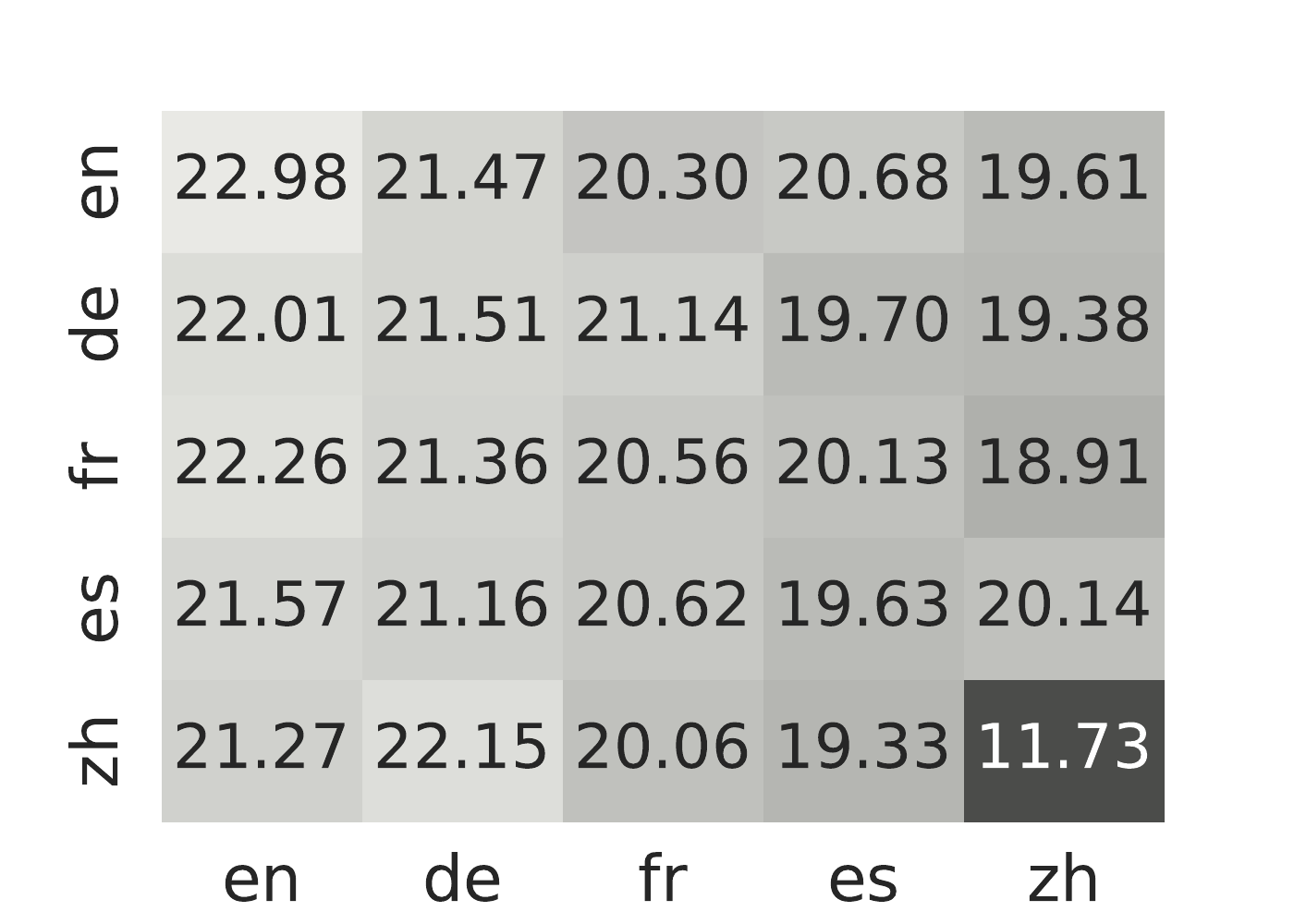}
  }\hspace{-2mm}
  \subfigure[Summ mBART]{
  \label{fig:Summ mBART}
  \includegraphics[width=0.17\linewidth]{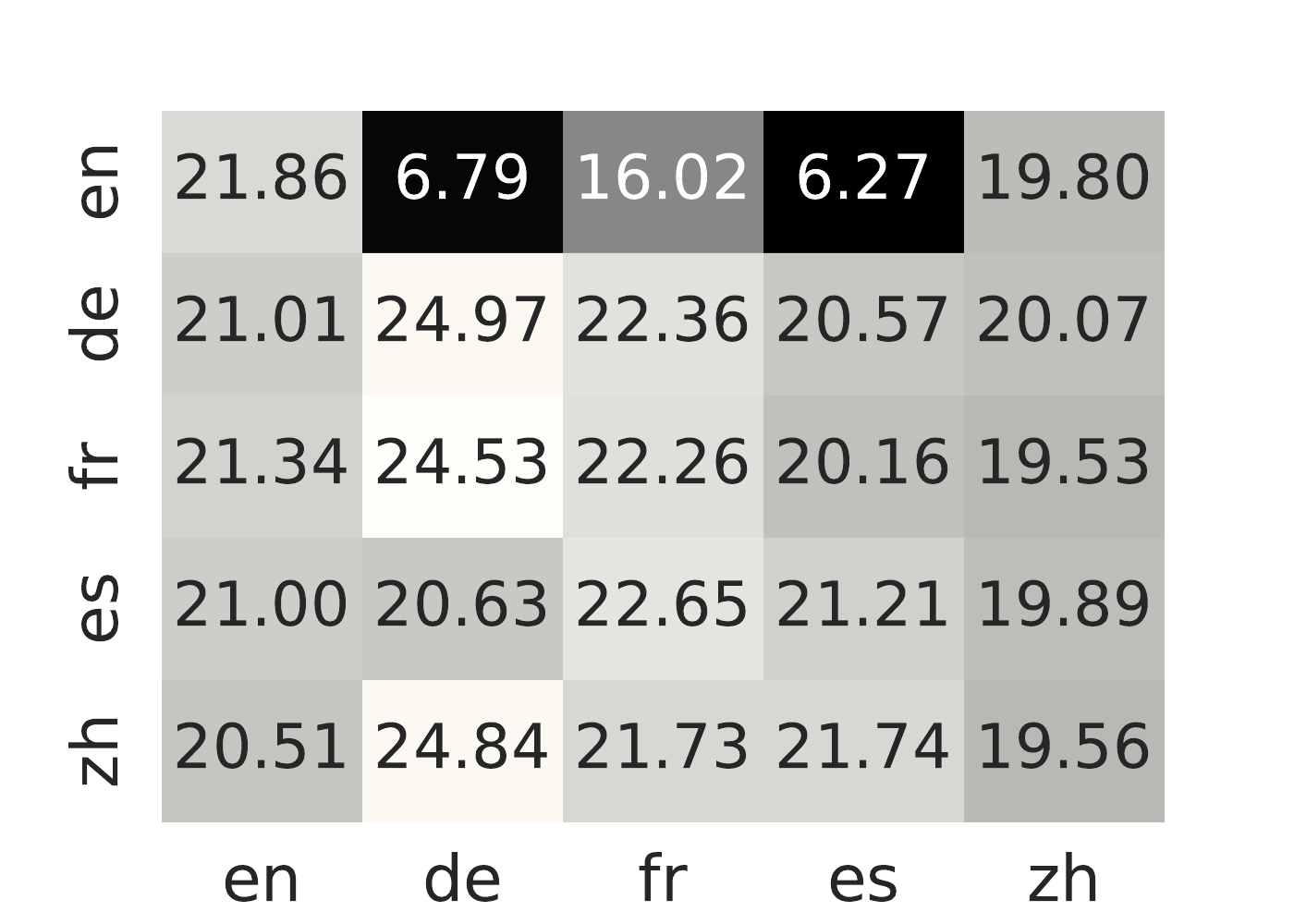}
  }\hspace{-2mm}
  \subfigure[Summ mT5]{
  \label{fig:Summ mT5}
  \includegraphics[width=0.18\linewidth]{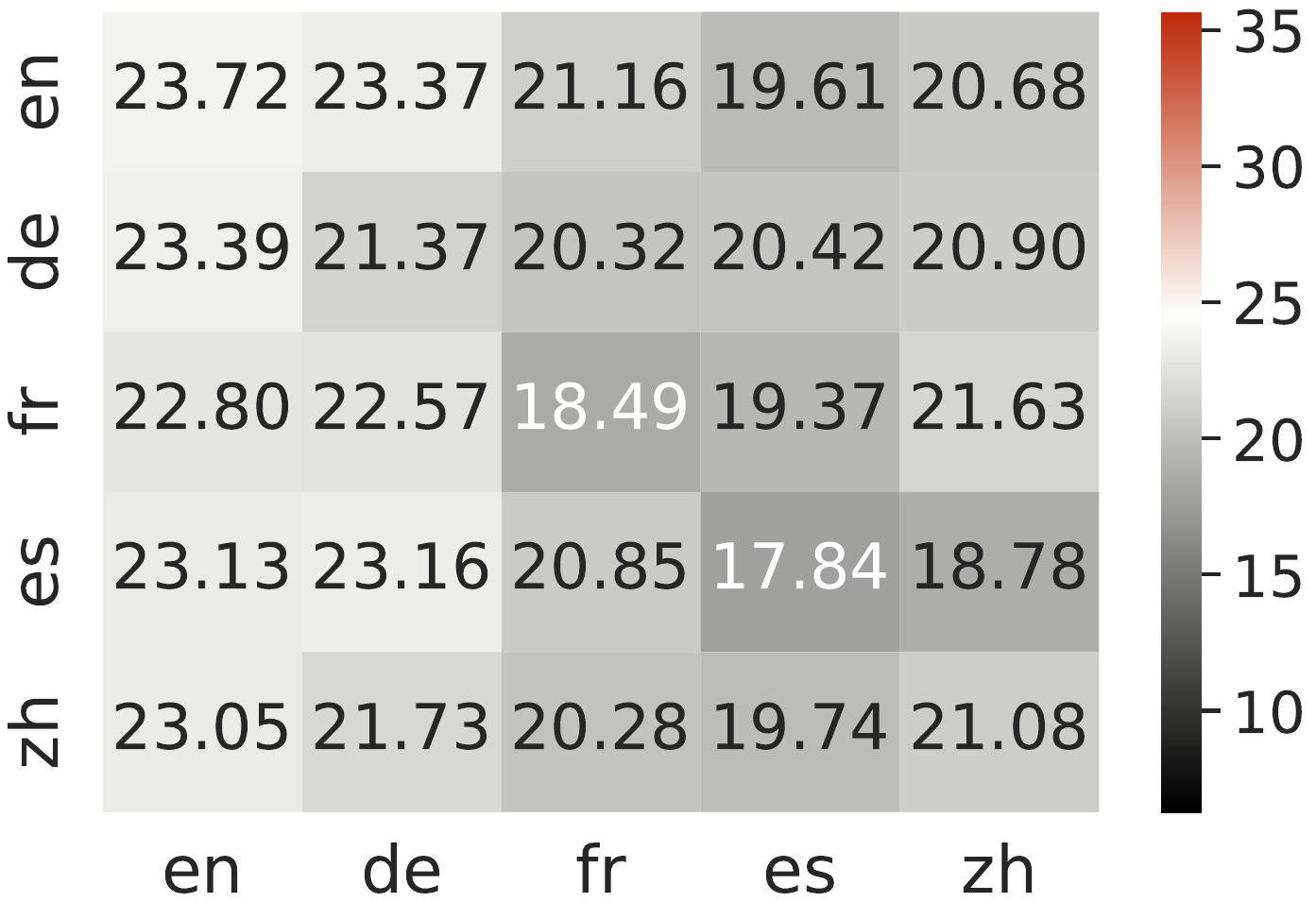}
  }
  \caption{The cross-lingual ensemble metric results for four models in four tasks. Every cell of row lang1 and column lang2 means the result when the languages of input and output are lang1 and lang2 respectively. Deeper \textcolor{red}{red} represents better cross-lingual performance while deeper \textcolor{gray}{gray} indicates worse performance.}
  \label{fig:crosslingual}
\end{figure*}

\subsection{Baseline Models}
\label{sec:models}
The performance of the following four popular multilingual pretrained models is explored\footnote{Detailed descriptions for models are included in Appendix \ref{sec:experimental settings}.}:

\textbf{M-BERT} Multilingual BERT (M-BERT)~\cite{devlin2019bert} is a language model pretrained from monolingual corpora in $104$ languages using Masked Language Modeling (MLM) task. 

\textbf{XLM} The Cross-Lingual Language Model (XLM)~\cite{lample2019cross} is pretrained with Masked Language Modeling (MLM) task using monolingual data and Translation Language Modeling (TLM) task using parallel data.

\textbf{mBART} Multilingual BART (mBART)~\cite{liu2020multilingual} is a pretrained encoder-decoder model using denoising auto-encoding objective on monolingual data over $25$ languages.

\textbf{mT$5$} Multilingual T$5$ (mT$5$)~\cite{xue2020mt5} is a multilingual variant of T$5$~\cite{2020t5} formatting all tasks as text-to-text generation problems.
mT$5$ is pretrained on a span-corruption version of Masked Language Modeling objective over $101$ languages.


\subsection{Evaluation Metrics}
In order to fully understand the model performance, the quality of generated texts is evaluated from different aspects, including metrics measuring the relevance between outputs and references (e.g., BLEU, ROUGE, and BERTScore) and metrics measuring the diversity of the generated texts (e.g., Distinct). 
Moreover, we propose a new ensemble metric leveraging relevance metrics to measure how close the generated text is to human writing.
It not only has higher correlation scores with human judgments but also is capable of measuring model performances fairly between languages.

\textbf{N-gram based Metrics}
N-gram based metrics evaluate the text-overlapping scores between the outputs and references. The following three metrics are used:
(1) \textbf{BLEU}~\cite{papineni2002bleu} is a popular metric that calculates the word-overlap scores between the generated texts and gold-standard ones. 
We use the BLEU-4, which is the average score for unigram, bigram, trigram, and 4-gram.
(2) \textbf{ROUGE}~\cite{lin2004rouge} is a recall-oriented metric that counts the number of overlapping units such as n-gram and word sequences between the produced texts and gold-standard ones. 
(3) \textbf{METEOR}~\cite{banerjee2005meteor} relies on semantic features to 
predict the similarity scores between system hypotheses and human references.

\begin{table*}[htbp]
  \centering
  \footnotesize
  \resizebox{0.99\textwidth}{!}{
    \begin{tabular}{clcccccccccccc}
    \toprule
    \multirow{2}[4]{*}{Task} & \multicolumn{1}{c}{\multirow{2}[4]{*}{Model}} & \multicolumn{2}{c}{BLEU} & \multicolumn{2}{c}{ROUGE-L} & \multicolumn{2}{c}{METEOR} & \multicolumn{2}{c}{BERTScore} & \multicolumn{2}{c}{Distinct-1} & \multicolumn{2}{c}{Ensemble} \\
\cmidrule{3-14}          &       & mono  & multi & mono  & multi & mono  & multi & mono  & multi & mono  & multi & mono  & multi \\
    \midrule
    \multirow{4}[2]{*}{SG} & M-BERT & 2.486  & \cellcolor[rgb]{ .906,  .902,  .902}\textbf{2.836 } & 16.680  & \cellcolor[rgb]{ .906,  .902,  .902}\textbf{17.240 } & 0.139  & \cellcolor[rgb]{ .906,  .902,  .902}\textbf{0.140 } & 0.741  & \cellcolor[rgb]{ .906,  .902,  .902}\textbf{0.743 } & 0.952  & \cellcolor[rgb]{ .906,  .902,  .902}\textbf{0.959 } & 30.891  & \cellcolor[rgb]{ .906,  .902,  .902}\textbf{30.987 } \\
          & XLM   & \textbf{4.026 } & \cellcolor[rgb]{ .906,  .902,  .902}2.992  & \textbf{24.520 } & \cellcolor[rgb]{ .906,  .902,  .902}22.820  & \textbf{0.145 } & \cellcolor[rgb]{ .906,  .902,  .902}0.144  & \textbf{0.754 } & \cellcolor[rgb]{ .906,  .902,  .902}0.744  & \textbf{0.967 } & \cellcolor[rgb]{ .906,  .902,  .902}0.967  & 28.364  & \cellcolor[rgb]{ .906,  .902,  .902}\textbf{28.449 } \\
          & mBART & 4.514  & \cellcolor[rgb]{ .906,  .902,  .902}\textbf{4.880 } & 19.320  & \cellcolor[rgb]{ .906,  .902,  .902}\textbf{19.920 } & 0.149  & \cellcolor[rgb]{ .906,  .902,  .902}\textbf{0.156 } & 0.759  & \cellcolor[rgb]{ .906,  .902,  .902}\textbf{0.762 } & \textbf{0.985 } & \cellcolor[rgb]{ .906,  .902,  .902}0.983  & 31.430  & \cellcolor[rgb]{ .906,  .902,  .902}\textbf{31.907 } \\
          & mT5   & 2.668  & \cellcolor[rgb]{ .906,  .902,  .902}\textbf{3.832 } & 16.280  & \cellcolor[rgb]{ .906,  .902,  .902}\textbf{18.620 } & 0.126  & \cellcolor[rgb]{ .906,  .902,  .902}\textbf{0.145 } & 0.751  & \cellcolor[rgb]{ .906,  .902,  .902}\textbf{0.759 } & \textbf{0.976 } & \cellcolor[rgb]{ .906,  .902,  .902}0.974  & \textbf{31.623 } & \cellcolor[rgb]{ .906,  .902,  .902}31.482  \\
    \midrule
    \multirow{4}[2]{*}{QG} & M-BERT & 8.266  & \cellcolor[rgb]{ .906,  .902,  .902}\textbf{9.980 } & 27.340  & \cellcolor[rgb]{ .906,  .902,  .902}\textbf{29.520 } & 0.240  & \cellcolor[rgb]{ .906,  .902,  .902}\textbf{0.262 } & 0.778  & \cellcolor[rgb]{ .906,  .902,  .902}\textbf{0.785 } & 0.938  & \cellcolor[rgb]{ .906,  .902,  .902}\textbf{0.944 } & 30.553  & \cellcolor[rgb]{ .906,  .902,  .902}30.526  \\
          & XLM   & \textbf{16.472 } & \cellcolor[rgb]{ .906,  .902,  .902}15.264  & \textbf{41.100 } & \cellcolor[rgb]{ .906,  .902,  .902}40.600  & \textbf{0.305 } & \cellcolor[rgb]{ .906,  .902,  .902}0.298  & \textbf{0.810 } & \cellcolor[rgb]{ .906,  .902,  .902}0.809  & 0.966  & \cellcolor[rgb]{ .906,  .902,  .902}\textbf{0.967 } & \textbf{30.072 } & \cellcolor[rgb]{ .906,  .902,  .902}29.979  \\
          & mBART & 16.256  & \cellcolor[rgb]{ .906,  .902,  .902}\textbf{17.624 } & 36.640  & \cellcolor[rgb]{ .906,  .902,  .902}\textbf{38.140 } & 0.298  & \cellcolor[rgb]{ .906,  .902,  .902}\textbf{0.315 } & 0.811  & \cellcolor[rgb]{ .906,  .902,  .902}\textbf{0.817 } & 0.981  & \cellcolor[rgb]{ .906,  .902,  .902}\textbf{0.983 } & 32.522  & \cellcolor[rgb]{ .906,  .902,  .902}\textbf{32.961 } \\
          & mT5   & 15.792  & \cellcolor[rgb]{ .906,  .902,  .902}\textbf{17.700 } & 34.100  & \cellcolor[rgb]{ .906,  .902,  .902}\textbf{37.680 } & 0.294  & \cellcolor[rgb]{ .906,  .902,  .902}\textbf{0.313 } & 0.806  & \cellcolor[rgb]{ .906,  .902,  .902}\textbf{0.818 } & 0.977  & \cellcolor[rgb]{ .906,  .902,  .902}\textbf{0.979 } & 32.257  & \cellcolor[rgb]{ .906,  .902,  .902}\textbf{32.944 } \\
    \midrule
    \multirow{4}[2]{*}{TG} & M-BERT & 9.524  & \cellcolor[rgb]{ .906,  .902,  .902}\textbf{10.550 } & 25.440  & \cellcolor[rgb]{ .906,  .902,  .902}\textbf{26.360 } & 0.214  & \cellcolor[rgb]{ .906,  .902,  .902}\textbf{0.228 } & 0.749  & \cellcolor[rgb]{ .906,  .902,  .902}\textbf{0.754 } & 0.930  & \cellcolor[rgb]{ .906,  .902,  .902}\textbf{0.957 } & 28.971  & \cellcolor[rgb]{ .906,  .902,  .902}\textbf{29.422 } \\
          & XLM   & 11.144  & \cellcolor[rgb]{ .906,  .902,  .902}\textbf{11.926 } & 26.960  & \cellcolor[rgb]{ .906,  .902,  .902}\textbf{28.660 } & 0.236  & \cellcolor[rgb]{ .906,  .902,  .902}\textbf{0.248 } & 0.752  & \cellcolor[rgb]{ .906,  .902,  .902}\textbf{0.759 } & \textbf{0.946 } & \cellcolor[rgb]{ .906,  .902,  .902}0.941  & 28.808  & \cellcolor[rgb]{ .906,  .902,  .902}\textbf{29.063 } \\
          & mBART & 14.726  & \cellcolor[rgb]{ .906,  .902,  .902}\textbf{14.786 } & 31.680  & \cellcolor[rgb]{ .906,  .902,  .902}\textbf{32.120 } & 0.257  & \cellcolor[rgb]{ .906,  .902,  .902}\textbf{0.260 } & 0.773  & \cellcolor[rgb]{ .906,  .902,  .902}\textbf{0.775 } & 0.966  & \cellcolor[rgb]{ .906,  .902,  .902}\textbf{0.968 } & \textbf{30.556 } & \cellcolor[rgb]{ .906,  .902,  .902}30.322  \\
          & mT5   & 11.336  & \cellcolor[rgb]{ .906,  .902,  .902}\textbf{13.546 } & 26.460  & \cellcolor[rgb]{ .906,  .902,  .902}\textbf{29.400 } & 0.223  & \cellcolor[rgb]{ .906,  .902,  .902}\textbf{0.257 } & 0.753  & \cellcolor[rgb]{ .906,  .902,  .902}\textbf{0.767 } & \textbf{0.959 } & \cellcolor[rgb]{ .906,  .902,  .902}0.956  & 29.556  & \cellcolor[rgb]{ .906,  .902,  .902}\textbf{30.010 } \\
    \midrule
    \multirow{4}[2]{*}{Summ} & M-BERT & 9.766  & \cellcolor[rgb]{ .906,  .902,  .902}\textbf{10.956 } & 31.280  & \cellcolor[rgb]{ .906,  .902,  .902}\textbf{32.220 } & 0.221  & \cellcolor[rgb]{ .906,  .902,  .902}\textbf{0.232 } & 0.748  & \cellcolor[rgb]{ .906,  .902,  .902}\textbf{0.751 } & 0.787  & \cellcolor[rgb]{ .906,  .902,  .902}\textbf{0.815 } & \textbf{22.122 } & \cellcolor[rgb]{ .906,  .902,  .902}22.018  \\
          & XLM   & 9.486  & \cellcolor[rgb]{ .906,  .902,  .902}\textbf{11.830 } & 30.160  & \cellcolor[rgb]{ .906,  .902,  .902}\textbf{34.740 } & \textbf{0.235 } & \cellcolor[rgb]{ .906,  .902,  .902}0.235  & 0.729  & \cellcolor[rgb]{ .906,  .902,  .902}\textbf{0.755 } & \textbf{0.814 } & \cellcolor[rgb]{ .906,  .902,  .902}0.772  & 19.281  & \cellcolor[rgb]{ .906,  .902,  .902}\textbf{20.770 } \\
          & mBART & \textbf{12.858 } & \cellcolor[rgb]{ .906,  .902,  .902}12.792  & \textbf{32.940 } & \cellcolor[rgb]{ .906,  .902,  .902}32.920  & 0.256  & \cellcolor[rgb]{ .906,  .902,  .902}\textbf{0.257 } & 0.750  & \cellcolor[rgb]{ .906,  .902,  .902}\textbf{0.750 } & 0.796  & \cellcolor[rgb]{ .906,  .902,  .902}\textbf{0.803 } & 21.972  & \cellcolor[rgb]{ .906,  .902,  .902}\textbf{22.292 } \\
          & mT5   & 5.022  & \cellcolor[rgb]{ .906,  .902,  .902}\textbf{6.090 } & 25.060  & \cellcolor[rgb]{ .906,  .902,  .902}\textbf{27.980 } & 0.145  & \cellcolor[rgb]{ .906,  .902,  .902}\textbf{0.162 } & 0.724  & \cellcolor[rgb]{ .906,  .902,  .902}\textbf{0.741 } & 0.826  & \cellcolor[rgb]{ .906,  .902,  .902}\textbf{0.870 } & 20.499  & \cellcolor[rgb]{ .906,  .902,  .902}\textbf{21.826 } \\
    \bottomrule
    \end{tabular}%
    }
\caption{Automatic scores averaged across five languages for four models on four tasks. Mono and multi mean models are trained in monolingual and multilingual setting respectively. Higher scores between monolingual and multilingual results are bolded.}
\label{tab:mono vs multi}%
\end{table*}%

\textbf{Embedding based Metrics}
The embedding-based metrics can, to a large extent, capture the semantic-level similarity between the generated text and the ground truth.
\textbf{BERTScore}~\cite{zhang2019bertscore} computes the similarity of candidate and reference as a sum of cosine similarities of tokens using BERT contextual embeddings.

\textbf{Diversity Metrics}
We also employ the distinct metric~\cite{li2016diversity}, which calculates the proportion of the distinct n-grams in all the system hypotheses and can be used to evaluate the diversity of the generated texts. 

\textbf{Human Evaluation}
Human evaluation is also leveraged to better estimate the quality of model outputs. Specifically, $30$ cases are randomly sampled from the test set for each task and language while ensuring all $30$ cases are aligned among five languages, and then they are presented to human annotators with the model outputs. The generated texts are evaluated under task-agnostic and task-specific aspects. Task-agnostic aspects include \textbf{Grammar}, \textbf{Fluency}, \textbf{Relevance} and \textbf{Language Fusion}. The former three aspects are scored from $1$ to $5$ while the language fusion score is set to $1$ if all tokens of a model-generated text are in the target language and $0$ otherwise. 

Besides task-agnostic aspects, the generated text is also evaluated under task-specific aspects. For title generation and summarization, coverage measures the degree to which the generated text covers the main content of the document. Correspondence for question generation measures the extent to which the generated question is matched with both document and answer. For story generation, we further evaluate whether the generated story is logically feasible. All task-specific aspects are scored from $1$ to $5$. 

\textbf{Ensemble Metric}
Some N-gram based metrics such as BLEU and ROUGE largely depend on the tokenizer for specific languages. For example, BLEU scores for Chinese outputs are relatively high because it simply uses a character-level tokenizer. This causes unfair comparison between different languages. To this end, we propose an ensemble metric that evaluates the degree to which a piece of text resembles manual writing. It not only enables fair comparison between languages but is also proved to have a better correlation with human-annotated scores at the end of this subsection.
We first average the grammar, fluency and relevance scores as targets, then normalize the automatic metrics and human scores among every language to eliminate the score discrepancy between languages. Three relevance metrics (BLEU, ROUGE-L, and BERTScore-F1) are gathered as features. The samples are split into training, development and test sets. 

After comparing different regression models' performance as shown in the upper part of Table \ref{tab:correlation all}
, we finally choose bagging regression model~\cite{breiman1996bagging} as the ensemble metric. 
Moreover, the bagging ensemble metric shows a higher correlation with human-annotated scores compared with other relevance automatic metrics as shown in the lower part of Table \ref{tab:correlation all}.

\subsection{Evaluation Scenarios}
\label{sec:scenarios}
To validate the effect of different experimental settings on model performance, several state-of-the-art multilingual models are studied under four evaluation scenarios.


\textbf{Monolingual fine-tuning} The pretrained model is tuned for a downstream task using the training data for a specific language and evaluated on the test set for the same language. 


\textbf{Multilingual fine-tuning} The pretrained model is jointly fine-tuned with data in all languages for a specific task. 
Different from the monolingual fine-tuning setting, there is only one model for each downstream task, which can serve all languages. 


\textbf{Cross-lingual generation} Since \dataset is multiway parallel, it can be reorganized to create input-output pairs that belong to different languages. 
In this paper, we make use of the multiway parallel data to do the supervised cross-lingual training, e.g., for English centric cross-lingual training, we take the English source as the input and the parallel German, French, Spanish, Chinese target as the output. Then we evaluate the model on same setting (en->de, en->es, en->fr, en->zh).
The cross-lingual generation performances on all $5*4$ directions are evaluated.

\textbf{Zero-shot transfer} We also try to explore the zero-shot ability of multilingual pretrained models on the four tasks. The model is fine-tuned on a specific task with English input and output. Then it is used to generate output in other languages with a given language tag.

\begin{table*}[htbp]
  \centering
  \footnotesize
  \resizebox{0.95\textwidth}{!}{
    \begin{tabular}{clccccccccc}
    \toprule
    \textbf{Task} & \multicolumn{1}{c}{\textbf{Language}} & \textbf{BLEU} & \textbf{ROUGE-1} & \textbf{ROUGE-2} & \textbf{ROUGE-L} & \textbf{METEOR} & \textbf{BERTScore} & \textbf{Distinct-1} & \textbf{Distinct-2} & \textbf{Ensemble} \\
    \midrule
    \multirow{4}[2]{*}{SG} & en->de & 0.02/3.20 & 7.20/27.20 & 0.20/4.00 & 7.20/25.80 & 0.05/0.14 & 0.63/0.73 & 0.47/0.96 & 0.50/1.00 & 18.90/29.70 \\
          & en->fr & 0.02/4.23 & 5.90/28.10 & 0.20/6.30 & 5.90/26.40 & 0.04/0.20 & 0.63/0.74 & 0.38/0.95 & 0.41/0.99 & 14.30/27.70 \\
          & en->es & 0.09/3.38 & 8.70/26.30 & 0.40/4.60 & 8.50/24.80 & 0.04/0.14 & 0.65/0.74 & 0.52/0.96 & 0.55/0.99 & 16.90/28.40 \\
          & en->zh & 0.00/5.79 & 0.00/28.80 & 0.00/8.80 & 0.00/26.80 & -     & 0.45/0.67 & 0.61/0.99 & 0.57/0.34 & 16.60/26.70 \\
    \midrule
    \multirow{4}[2]{*}{QG} & en->de & 1.96/10.41 & 18.10/38.70 & 2.40/14.70 & 17.60/37.20 & 0.10/0.25 & 0.73/0.78 & 0.94/0.97 & 0.98/1.00 & 29.80/29.30 \\
          & en->fr & 2.16/14.70 & 16.80/42.80 & 2.90/19.00 & 16.20/39.60 & 0.08/0.35 & 0.74/0.80 & 0.94/0.95 & 0.99/0.99 & 28.60/29.80 \\
          & en->es & 7.46/16.93 & 25.50/49.50 & 8.70/22.40 & 23.90/46.80 & 0.18/0.37 & 0.76/0.83 & 0.94/0.95 & 0.99/1.00 & 28.50/29.10 \\
          & en->zh & 0.00/16.07 & 0.00/43.10 & 0.00/22.90 & 0.00/37.90 & -     & 0.44/0.73 & 0.10/1.00 & 0.08/1.00 & 16.40/28.60 \\
    \midrule
    \multirow{4}[2]{*}{TG} & en->de & 2.58/9.15 & 13.40/26.90 & 4.40/11.10 & 12.50/24.30 & 0.12/0.22 & 0.67/0.73 & 0.83/0.95 & 0.88/0.99 & 26.30/30.60 \\
          & en->fr & 3.26/11.54 & 13.90/33.80 & 4.50/14.70 & 12.70/29.00 & 0.12/0.30 & 0.69/0.75 & 0.89/0.91 & 0.93/0.99 & 25.20/28.50 \\
          & en->es & 4.90/12.45 & 21.20/36.30 & 7.40/15.70 & 18.50/31.10 & 0.17/0.31 & 0.71/0.76 & 0.88/0.91 & 0.94/0.99 & 24.50/29.50 \\
          & en->zh & 0.01/15.44 & 0.00/34.50 & 0.00/19.40 & 0.00/29.90 & -     & 0.45/0.69 & 0.37/0.98 & 0.22/0.58 & 16.70/27.10 \\
    \midrule
    \multirow{4}[2]{*}{Summ} & en->de & 1.85/8.36 & 15.40/34.70 & 2.90/11.70 & 14.50/31.10 & 0.08/0.20 & 0.65/0.72 & 0.61/0.81 & 0.78/0.97 & 18.50/21.50 \\
          & en->fr & 1.29/11.79 & 13.70/39.90 & 2.60/15.80 & 13.00/35.50 & 0.07/0.29 & 0.68/0.75 & 0.64/0.75 & 0.82/0.94 & 18.60/20.30 \\
          & en->es & 4.18/11.93 & 22.50/41.00 & 5.80/15.60 & 20.30/36.60 & 0.14/0.29 & 0.69/0.75 & 0.64/0.74 & 0.82/0.95 & 17.30/20.70 \\
          & en->zh & 0.00/14.58 & 0.00/42.20 & 0.00/20.40 & 0.00/38.70 & -     & 0.42/0.71 & 0.68/0.84 & 0.27/0.94 & 12.80/19.60 \\
    \bottomrule
    \end{tabular}%
    }
  \caption{English centric zero-shot and cross-lingual results for XLM on four tasks. Scores on the left and right side of each cell represent the zero-shot and cross-lingual results respectively.}
  \label{tab:zeroshot_vs_crosslingual}%
\end{table*}%

\section{Results}

\subsection{Monolingual and Cross-lingual}
This section displays the monolingual and cross-lingual model comparison to explore their performances in different tasks and languages. 
Figure \ref{fig:crosslingual} contains the five language-centric cross-lingual and monolingual results. Several conclusions can be drawn from the results: 

\textbf{The performance of Cross-lingual is better than monolingual in some cases. }
As shown in Figure \ref{fig:crosslingual}, model performances on ensemble scores in cross-lingual setting exceed those in monolingual setting frequently (e.g., the monolingual result of French underperforms the English to French cross-lingual result in Figure \ref{fig:SG XLM} ). This is because 
the cross-lingual models are trained with more data (e.g., the English centric cross-lingual model is trained with en->de, en->fr, en->es, en->zh data), and the data from different cross-lingual directions can sometimes benefit from each other thus improving the model performance. 

\textbf{Chinese text generation is challenging in cross-lingual setting.} As illustrated in Figure \ref{fig:crosslingual}, nearly all models obtain inferior scores when generating Chinese text. Also, model results on Chinese inputs are usually worse than results on inputs in other languages.
The wide discrepancies in grammar and vocabulary between Chinese and other languages lead to the poor performance of cross-lingual generation when either the target language or source language is Chinese. 

\textbf{Multilingual pretrained models obtain lower scores on the Summarization task.} Compared with other tasks, summarization task requires longer output, which increases the difficulty of text generation, thus causing poor performance both in cross-lingual and monolingual settings.

\subsection{Monolingual and Multilingual}
\label{sec: mono vs multi}
In addition to cross-lingual analysis, we also explore the performance difference between models trained in monolingual and multilingual settings. Table \ref{tab:mono vs multi} displays the monolingual and multilingual training results for four models in four tasks.

\textbf{In most cases, multilingual training can improve model performance on relevance.} As shown in Table \ref{tab:mono vs multi}, $75$ out of $96$ multilingual results outperform the monolingual counterparts on various relevance metrics in different tasks. The reason is that the multilingual data in \dataset is fully parallel across all five languages and every sample has semantically aligned counterparts in other languages.
It makes better semantic fusion among different languages, thus boosting the multilingual training performance.

\textbf{The advantages of multilingual training are not obvious on diversity measured by distinct-$1$.}
Especially in the story generation task, $3$ out of $4$ models obtain better distinct-1 scores in monolingual setting than in multilingual one. 
Diversity can not be improved by semantic sharing across languages especially when the samples of them are multiway parallel. This is because the multiway parallel dataset with the semantic aligned samples repeating in different languages encourages models to generate similar texts to some extent.

\begin{figure}
  \centering
  \includegraphics[width=0.75\linewidth]{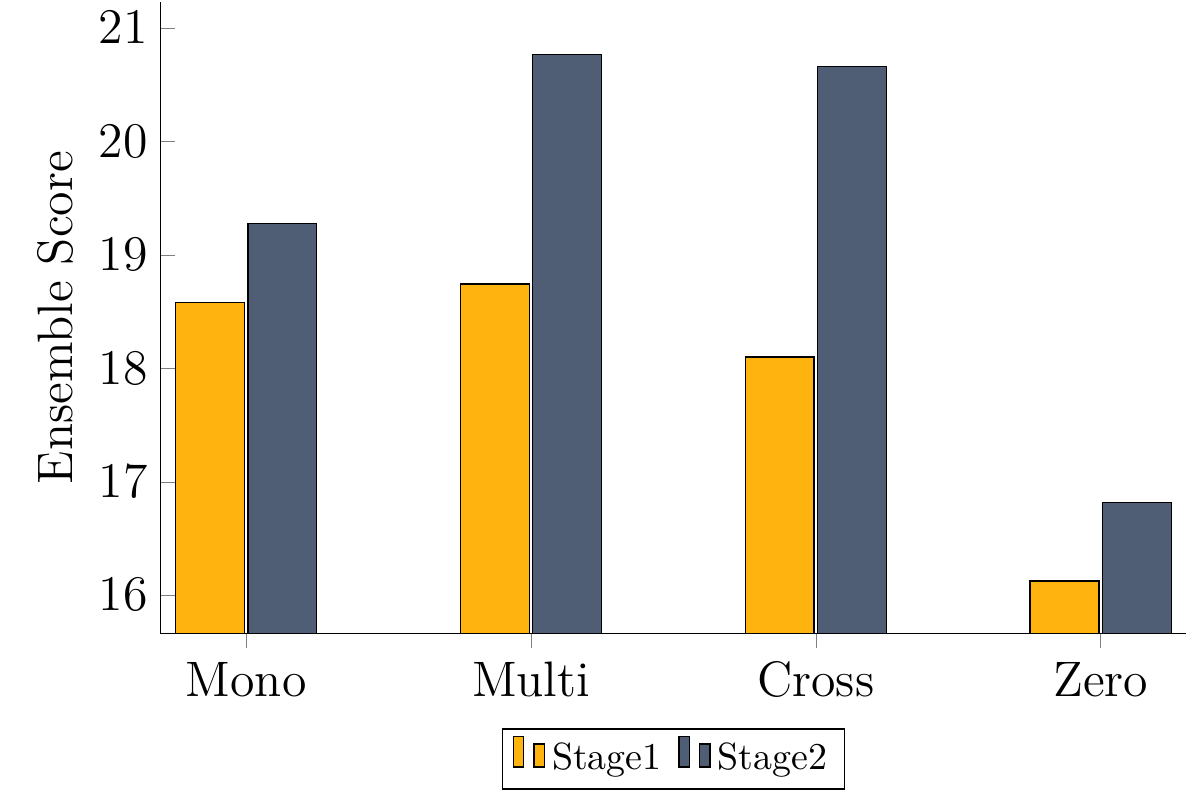}
  \caption{The different stage performances averaged across five languages of XLM in summarization under various settings. Here stage1 represents models trained only on rough training data while stage2 represents models further trained on human-annotated training data based on models in stage1.}
  \label{fig:diss1}
\end{figure}

\subsection{Zero-shot results}
To test the cross-lingual generation ability of multilingual pretrained models when no direct cross-lingual training data are provided, we evaluate the zero-shot cross-lingual generation performance.

Table \ref{tab:zeroshot_vs_crosslingual} presents the zero-shot results for XLM in four tasks. It demonstrates that 
the multilingual pretrained model XLM still lacks the ability to generate high-quality cross-lingual output in zero-shot scenario. \textbf{Moreover, English to Chinese and French zero-shot generation shows inferior performance.}\footnote{Zero-shot results show the same trend as shown in Table \ref{tab:zero all} in Appendix.} The performance decline is rather salient when generating Chinese text. This is because Chinese and French (especially Chinese) are distant from English in the language family tree. \textbf{On the other hand, zero-shot results underperform cross-lingual results} which further emphasizes the importance of direct cross-lingual training data for cross-lingual text generation.

\subsection{Pseudo and Annotated Data}
\textbf{To answer the question ``Does the 400k annotated training data help the model generate better? '',}
we use the rough training data filtered by back translation for the first stage fine-tuning and the annotated training data for the second stage.  The ablation study results on the two-step fine-tuning in summarization under all evaluation scenarios with XLM are illustrated in Figure \ref{fig:diss1}. 

The extra human-annotated data boost model performance by at least $3.8\%$ on the ensemble metric.
We also make a T-test and prove that the improvement of annotated training data is significant in all settings.\footnote{The t-test details are shown in Appendix~\ref{sec:stag1 vs stage2 significant test}.} \textbf{It demonstrates that although the number of annotated data is small, it can significantly improve the performance.} It also highlights the necessity of human-annotated multilingual data compared with pseudo-parallel data via machine translation.


\begin{table}[tbp]
  \footnotesize
  \renewcommand\arraystretch{0.7}
  \centering
  \resizebox{0.47\textwidth}{!}{
    \begin{tabular}{clccccc}
    \toprule
    \multicolumn{1}{l}{\textbf{Setting}} & \textbf{Model} & \textbf{Gram.} & \textbf{Flu.} & \textbf{Rel.} & \textbf{lang fuse} & \textbf{task spec.} \\
    \midrule
    \multirow{4}[2]{*}{SG} & mono & 4.69  & 4.81  & 3.75  & 1.00  & 3.79  \\
          & multi & 4.71  & 4.80  & 3.67  & 1.00  & 4.02  \\
          & cross & 4.18  & 4.23  & 3.49  & 0.95  & 2.53  \\
          & zero & 4.15  & 4.18  & 3.27  & 0.18  & 3.00  \\
    \midrule
    \multirow{4}[2]{*}{QG} & mono & 4.66  & 4.69  & 3.03  & 0.99  & 3.95  \\
          & multi & 4.69  & 4.67  & 3.06  & 0.97  & 4.11  \\
          & cross & 4.30  & 4.30  & 2.70  & 0.95  & 2.64  \\
          & zero & 3.35  & 4.26  & 3.18  & 0.19  & 3.09  \\
    \midrule
    \multirow{4}[2]{*}{TG} & mono & 4.53  & 4.51  & 3.09  & 0.96  & 3.71  \\
          & multi & 4.66  & 4.65  & 3.18  & 0.93  & 3.17  \\
          & cross & 3.73  & 3.64  & 2.63  & 0.90  & 1.85  \\
          & zero & 3.52  & 4.15  & 3.51  & 0.18  & 1.43  \\
    \midrule
    \multirow{4}[2]{*}{Summ} & mono & 4.19  & 3.99  & 3.71  & 0.68  & 3.71  \\
          & multi & 4.19  & 4.02  & 3.78  & 0.64  & 3.60  \\
          & cross & 2.14  & 2.22  & 2.23  & 0.68  & 2.05  \\
          & zero & 1.57  & 1.54  & 1.58  & 0.03  & 1.59  \\
    \bottomrule
    \end{tabular}%
    }
  \caption{Human evaluation scores averaged on five languages for mBART on four tasks. `Gram.', `Flu.', `Rel.', `Lang Fu.', `Task Spec.' indicates \textbf{Grammar}, \textbf{Fluency}, \textbf{Relevance}, \textbf{Language Fusion} and \textbf{Task Specific} scores respectively. 
  }
  \label{tab:human}%
\end{table}%

\subsection{Human evaluation}
Table \ref{tab:human} presents the human evaluation scores for mBART in four tasks. Multilingual training results can surpass the monolingual results in QG, TG and Summ on relevance. In terms of task-specific score, multilingual results are also superior in SG and QG. This is consistent with the conclusion in Sec. \ref{sec: mono vs multi}. On the other hand, language fusion scores 
in zero-shot setting are extremely low, indicating the pretrained models still lack the ability to generate texts in correct language in zero-shot setting. 

\section{Discussions}

Considering the annotation cost, it is not realistic to construct a multiway text generation dataset with all data annotated by human. As a consequence, most of the non-English data in MTG are automatically translated from their English counterparts. Although the n-gram consistency check when round-trip translating the data can guarantee the quality of them to some extent, some translation errors are inevitable. MTG with more annotated data and with data filtered by more reliable methods will be explored in the future.

On the other hand, human often gives an overall evaluation of a generated text rather than measuring it in fine-grained aspects of grammar, fluency and relevance. Thus we try to propose a metric measuring how a text resembles human writing and consider grammar, fluency and relevance as a whole. This metric may not be perfect, but it is a promising direction as there does not exist a really reliable text generation metric nowadays.

\section{Conclusion}

In this paper, we propose a multilingual multiway benchmark \dataset for text generation. It contains four typical generation tasks: story, question, title generation and text summarization. The key feature of \dataset is that it has multiway parallel data across five diverse languages: English, German, French, Spanish and Chinese. 
It provides the benchmark with the ability to create cross-lingual data between any two languages and makes the semantic fusion between languages easier. 
On the other hand, it provides more evaluation scenarios, such as multilingual training, cross-lingual generation and zero-shot transfer. 
We also benchmark state-of-the-art multilingual pretrained models on our \dataset from different metrics (including a newly proposed ensemble metric) to explore its features and promote research in multilingual text generation.

\section{Ethics Consideration}
Since we propose a new multilingual text generation benchmark \dataset, we solve some possible ethic considerations in this section. 
\paragraph{English dataset}
We choose ROCStories, SQUAD 1.0, ByteCup
and CNN/DailyMail as the English datasets for story, question, title generation and text summarization tasks. All of them are available for research use under their licenses. They can be downloaded free from their websites\footnote{ROCStories requires for some necessary contact information}.
We ensure that these datasets are only used for academic research and the dataset construction process complies with the intellectual property
and privacy rights of the original authors. Also, our proposed benchmark suite \dataset should only be used for academic research purposes.

\paragraph{Annotation process}
As described in Sec. \ref{sec: data collection}, we hire some full-time and part-time language experts to do the annotation. Full-time experts are paid $\$40$ per day and part-time annotators are paid $\$0.2$ per example\footnote{Full-time employees work at most $8$ hours per day, and the local minimum hourly wage is $\$3.7$. The part-time annotators can produce at least $20$ examples per hour.}. Their salary is higher than the local average hourly minimum wage. All annotators are aware of any risk of harm associated with their participation. The annotation process is in compliance with the intellectual property and privacy rights of the recruited annotators. The annotation protocol is proved by the legal department inside the company.

\paragraph{Risk Concern}
In this paper we propose a new ensemble metric measuring to what degree is the generated text close to human-level. The further pursue for more human-like multilingual generation will possibly raise safety concerns.



\bibliography{anthology}
\bibliographystyle{acl_natbib}

\appendix

\section{Back Translation Threshold Testing}
\label{sec:back translation threshold}
The detailed data sizes of back translation filtered datasets for different tasks are presented in Table \ref{tab:threshold testing}.


\section{Experimental settings}
\label{sec:experimental settings}
The overall statistics for multilingual pretrained models are presented in Table \ref{tab:model statistics} and the detailed descriptions for them are as follows:

1\textbf{M-BERT} Multilingual BERT (M-BERT)~\cite{devlin2019bert} is a single language model pre-trained from monolingual corpora in $104$ languages using Masked Language Modeling (MLM) task. M-BERT leverages a shared vocabulary of $110$k WordPiece tokens and has $12$ layers with $172$M parameters totally.

\textbf{XLM} The Cross-Lingual Language Model (XLM)~\cite{lample2019cross} is pre-trained simultaneously with Masked Language Model (MLM) task using monolingual data and Translation Language Model (TLM) task using parallel data. XLM has a shared vocabulary of $200$k byte-pair encoded (BPE) subwords~\cite{sennrich2016neural} and $16$ layers totaling $570$M parameters.

1\textbf{mBART} Multilingual BART (mBART)~\cite{liu2020multilingual} is a pre-trained encoder-decoder model using denoising auto-encoding objective on monolingual data over $25$ languages. mBART has a shared vocabulary of $250$k tokens leveraging Sentence Piece tokenization scheme. mBART consists of $12$-layer encoder and $12$-layer decoder with a total of $680$M parameters.

\textbf{mT$5$} Multilingual T$5$ (mT$5$)~\cite{xue2020mt5} is a multilingual variant of T$5$~\cite{2020t5} leveraging a text-to-text format. mT$5$ is pre-trained on a span-corruption version of Masked Language Modeling objective over $101$ languages. It is composed of $24$-encoder layers and $24$ decoder layers with $13$B parameters.

We use the encoder-decoder architecture for our generation tasks.
Among the models described above, mBART and mT5 have been pretrained for generation tasks, but M-BERT and XLM are only pretrained for encoder representations. Therefore, we initialize the decoder with the encoder parameters for M-BERT and XLM. 
During the pretraining phase, there are no language tags in M-BERT and mT5. Thus we manually add the language tag at the beginning of the source and target for M-BERT and add the target language tag to the beginning of source for mT5.

We adjust the input format for each task. For QG, we append the answer to the passage and insert a special token to separate them. For SG, we take the beginning four sentences as the source and make the last sentence as the target. 

We take a two-step finetuning to make full use of our \dataset benchmark. We first use the large rough parallel training data to train our models on the downstream tasks for 20 epochs, and then finetune the models on the small annotated training data to further improve the generation performance for 10 epochs. We evaluate the model for every 2000 steps and use the loss on development to choose the best model. The batch size is 32. The learning rate and optimizer parameters are set to the default parameters for each model. All models are trained with 32GB Tesla-V100.

\begin{table}[htbp]
\footnotesize
  \centering
    \begin{tabular}{ccccc}
    \toprule
    \textbf{Threshold} & \textbf{QG} & \textbf{TG} & \textbf{SG} & \textbf{Summ} \\
    \midrule
    0     & 82306 & 393792 & 88161 & 287083 \\
    0.3   & 80836 & 355034 & 88158 & 243698  \\
    0.4   & 79390 & 333461 & 88077 & 217777  \\
    0.5   & 71819 & 280376 & 87003 & 164355  \\
    0.6   & 32261 & 144109 & 75892 & 58060  \\
    \bottomrule
    \end{tabular}%
    \caption{The data sizes of datasets filtered by back translation with respect to different thresholds.}
  \label{tab:threshold testing}%
\end{table}%

\begin{table}[htbp]
  \footnotesize
  \centering
  \resizebox{0.48\textwidth}{!}{
    \begin{tabular}{lccccc}
    \toprule
    \textbf{Models} & \textbf{Arch} & \textbf{\# langs} & \textbf{\# vocab} & \textbf{\# layers} & \textbf{\# params} \\
    \midrule
    M-BERT & enc   & 104   & 110k  & 12    & 172M \\
    XLM   & enc   & 17    & 200k  & 16    & 570M \\
    mBART & enc-dec & 25    & 250k  & 12    & 680M \\
    mT5   & enc-dec & 101   & 250k  & 24    & 13B \\
    \bottomrule
    \end{tabular}%
    }
    \caption{The overall statistics for multilingual pretrained models. Arch means the architectures of models. \# vocab means the vocabulary sizes of models. \# langs, \# layers and \# params mean the number of languages, layers and parameters respectively.}
  \label{tab:model statistics}%
\end{table}%

\begin{table}[htbp]
  \centering
   \footnotesize
    \begin{tabular}{clcccc}
    \toprule
    \multicolumn{2}{c}{\textbf{Tasks}} & \textbf{Mono} & \textbf{Multi} & \textbf{Cross} & \textbf{Zero} \\
    \midrule
    \multirow{3}[2]{*}{SG} & stage1 & 0.268  & 0.270  & 0.258  & 0.125  \\
          & stage2 & \textbf{0.284 } & \textbf{0.284 } & \textbf{0.289 } & \textbf{0.167 } \\
          \cmidrule{2-6}
          & p-value & 0.000  & 0.000  & 0.000  & 0.000  \\
    \midrule
    \multirow{3}[2]{*}{QG} & stage1 & 0.286  & 0.287  & 0.279  & 0.235  \\
          & stage2 & \textbf{0.301 } & \textbf{0.300 } & \textbf{0.295 } & \textbf{0.258 } \\
          \cmidrule{2-6}
          & p-value & 0.000  & 0.000  & 0.000  & 0.000  \\
    \midrule
    \multirow{3}[2]{*}{TG} & stage1 & 0.257  & 0.270  & 0.268  & 0.223  \\
          & stage2 & \textbf{0.288 } & \textbf{0.291 } & \textbf{0.289 } & \textbf{0.232 } \\
          \cmidrule{2-6}
          & p-value & 0.000  & 0.000  & 0.000  & 0.003  \\
    \midrule
    \multirow{3}[2]{*}{Summ} & stage1 & 0.186  & 0.187  & 0.181  & 0.161  \\
          & stage2 & \textbf{0.193 } & \textbf{0.208 } & \textbf{0.207 } & \textbf{0.168 } \\
          \cmidrule{2-6}
          & p-value & 0.001  & 0.000  & 0.000  & 0.004  \\
    \bottomrule
    \end{tabular}%
    \caption{The average ensemble metric scores for XLM for stage1 and stage2 in four tasks in four settings and the corresponding t test p-values. Here stage1 represents models trained only on rouge training data while stage2 represents models further trained on human-annotated training data based on models in stage1. The bold cell means the significantly higher score between stage1 and stage2 scores.}
  \label{tab:significant test pvalue}%
\end{table}%

\section{Significant Test Results}
\label{sec:stag1 vs stage2 significant test}
The average ensemble metric scores for stage1 and stage2 in four tasks and the corresponding significant test p-values are displayed in Table \ref{tab:significant test pvalue}. As it shows, adding human-annotated training data can always improve the model performance under different settings. The improvements are significant in all settings.

\section{Experimental Results}
We present detailed experimental results of our four baseline models under four different evaluation settings here.

\begin{table*}[htbp]
  \renewcommand\arraystretch{0.6}
  \centering
\footnotesize
\resizebox{0.93\textwidth}{!}{
\begin{tabular}{cclccccccc}
    \toprule
    \multirow{2}[4]{*}{\textbf{Task}} & \multirow{2}[4]{*}{\textbf{Model}} & \multicolumn{1}{c}{\multirow{2}[4]{*}{\textbf{Language}}} & \multicolumn{3}{c}{\textbf{N-gram-based}} & \textbf{Embedding-based} & \multicolumn{2}{c}{\textbf{Diversity}} & \textbf{Ours} \\
\cmidrule{4-10}          &       &       & \textbf{BLEU} & \textbf{ROUGE-L} & \textbf{METEOR} & \textbf{BERTScore} & \textbf{Distinct-1} & \textbf{Distinct-2} & \textbf{Ensemble} \\
    \midrule
    \multirow{20}[8]{*}{SG} & \multirow{5}[2]{*}{M-BERT} & en->en & 2.56  & 18.8  & 0.103 & 0.894 & 0.917 & 0.99  & 31.254 \\
          &       & de->de & 2.27  & 13.4  & 0.131 & 0.714 & 0.944 & 0.994 & 34.812 \\
          &       & fr->fr & 1.38  & 12.7  & 0.201 & 0.715 & 0.945 & 0.997 & 31.209 \\
          &       & es->es & 1.81  & 13.5  & 0.121 & 0.72  & 0.955 & 0.996 & 30.825 \\
          &       & zh->zh & 4.41  & 25    & -     & 0.661 & 1     & 0.124 & 26.354 \\
\cmidrule{2-10}          & \multirow{5}[2]{*}{XLM} & en->en & 3.71  & 20.5  & 0.107 & 0.895 & 0.968 & 0.995 & 31.701 \\
\bottomrule
    \end{tabular}%
    }
  \caption{The whole results under the monolingual evaluation scenarios.}
  \label{tab:mono all all}%
\end{table*}%

\begin{table*}[htbp]
  \renewcommand\arraystretch{0.6}
  \centering
  \footnotesize
\resizebox{0.93\textwidth}{!}{
    \begin{tabular}{cclccccccc}
    \toprule
    \multirow{2}[4]{*}{\textbf{Task}} & \multirow{2}[4]{*}{\textbf{Model}} & \multicolumn{1}{c}{\multirow{2}[4]{*}{\textbf{Language}}} & \multicolumn{3}{c}{\textbf{N-gram-based}} & \textbf{Embedding-based} & \multicolumn{2}{c}{\textbf{Diversity}} & \textbf{Ours} \\
\cmidrule{4-10}          &       &       & \textbf{BLEU} & \textbf{ROUGE-L} & \textbf{METEOR} & \textbf{BERTScore} & \textbf{Distinct-1} & \textbf{Distinct-2} & \textbf{Ensemble} \\
    \midrule
    \multirow{20}[8]{*}{SG} & \multirow{5}[2]{*}{M-BERT} & en->en & 2.93  & 19.8  & 0.106 & 0.895 & 0.937 & 0.993 & 31.793 \\
          &       & de->de & 2.77  & 14.6  & 0.139 & 0.72  & 0.944 & 0.996 & 34.933 \\
          &       & fr->fr & 1.65  & 13.7  & 0.196 & 0.72  & 0.949 & 0.998 & 30.998 \\
          &       & es->es & 1.85  & 13.4  & 0.119 & 0.718 & 0.965 & 0.997 & 30.234 \\
          &       & zh->zh & 4.98  & 24.7  & -     & 0.661 & 0.998 & 0.374 & 26.977 \\
\cmidrule{2-10}          & \multirow{5}[2]{*}{XLM} & en->en & 3.56  & 20.2  & 0.106 & 0.896 & 0.966 & 0.997 & 31.594 \\
\bottomrule
    \end{tabular}%
    }
  \caption{The whole results under the multilingual evaluation scenarios.}
  \label{tab:multi all}%
\end{table*}%

\begin{table*}[htbp]
  \renewcommand\arraystretch{0.6}
  \centering
  \footnotesize
\resizebox{0.93\textwidth}{!}{
    \begin{tabular}{cclccccccc}
    \toprule
    \multirow{2}[4]{*}{\textbf{Task}} & \multirow{2}[4]{*}{\textbf{Model}} & \multicolumn{1}{c}{\multirow{2}[4]{*}{\textbf{Language}}} & \multicolumn{3}{c}{\textbf{N-gram-based}} & \textbf{Embedding-based} & \multicolumn{2}{c}{\textbf{Diversity}} & \textbf{Ours} \\
\cmidrule{4-10}          &       &       & \textbf{BLEU} & \textbf{ROUGE-L} & \textbf{METEOR} & \textbf{BERTScore} & \textbf{Distinct-1} & \textbf{Distinct-2} & \textbf{Ensemble} \\
    \midrule
    \multirow{16}[8]{*}{SG} & \multirow{4}[2]{*}{M-BERT} & en->de & 2.28  & 13.6  & 0.127 & 0.721 & 0.953 & 0.995 & 34.529 \\
          &       & en->fr & 1.43  & 14.1  & 0.182 & 0.72  & 0.938 & 0.995 & 30.507 \\
          &       & en->es & 1.57  & 13    & 0.111 & 0.72  & 0.943 & 0.993 & 30.368 \\
          &       & en->zh & 4.72  & 25.4  & -     & 0.664 & 0.997 & 0.302 & 26.423 \\
\cmidrule{2-10}          & \multirow{4}[2]{*}{XLM} & en->de & 3.2   & 25.8  & 0.142 & 0.73  & 0.964 & 0.995 & 29.683 \\
          &       & en->fr & 4.23  & 26.4  & 0.198 & 0.744 & 0.951 & 0.989 & 27.735 \\
\bottomrule
    \end{tabular}%
    }
  \caption{The whole results under the English centric cross-lingual evaluation scenarios.}
  \label{tab:cross en all}%
\end{table*}%

\begin{table*}[htbp]
  \renewcommand\arraystretch{0.6}
  \centering
  \footnotesize
\resizebox{0.93\textwidth}{!}{
    \begin{tabular}{cclccccccc}
    \toprule
    \multirow{2}[4]{*}{\textbf{Task}} & \multirow{2}[4]{*}{\textbf{Model}} & \multicolumn{1}{c}{\multirow{2}[4]{*}{\textbf{Language}}} & \multicolumn{3}{c}{\textbf{N-gram-based}} & \textbf{Embedding-based} & \multicolumn{2}{c}{\textbf{Diversity}} & \textbf{Ours} \\
\cmidrule{4-10}          &       &       & \textbf{BLEU} & \textbf{ROUGE-L} & \textbf{METEOR} & \textbf{BERTScore} & \textbf{Distinct-1} & \textbf{Distinct-2} & \textbf{Ensemble} \\
    \midrule
    \multirow{16}[8]{*}{SG} & \multirow{4}[2]{*}{M-BERT} & de->en & 2.59  & 18.5  & 0.1   & 0.893 & 0.918 & 0.988 & 30.794 \\
          &       & de->fr & 1.29  & 13.5  & 0.19  & 0.717 & 0.927 & 0.995 & 31.616 \\
          &       & de->es & 1.58  & 12.6  & 0.11  & 0.716 & 0.929 & 0.991 & 30.216 \\
          &       & de->zh & 4.57  & 24.1  & -     & 0.656 & 0.989 & 0.518 & 26.703 \\
\cmidrule{2-10}          & \multirow{4}[2]{*}{XLM} & de->en & 3.18  & 19.7  & 0.104 & 0.894 & 0.961 & 0.991 & 31.967 \\
          &       & de->fr & 3.85  & 26    & 0.189 & 0.744 & 0.949 & 0.986 & 27.691 \\
\bottomrule
    \end{tabular}%
    }
   \caption{The whole results under the German centric cross-lingual evaluation scenarios.}
  \label{tab:cross de all}%
\end{table*}%

\begin{table*}[htbp]
  \renewcommand\arraystretch{0.6}
  \centering
  \footnotesize
\resizebox{0.93\textwidth}{!}{
    \begin{tabular}{cclccccccc}
    \toprule
    \multirow{2}[4]{*}{\textbf{Task}} & \multirow{2}[4]{*}{\textbf{Model}} & \multicolumn{1}{c}{\multirow{2}[4]{*}{\textbf{Language}}} & \multicolumn{3}{c}{\textbf{N-gram-based}} & \textbf{Embedding-based} & \multicolumn{2}{c}{\textbf{Diversity}} & \textbf{Ours} \\
\cmidrule{4-10}          &       &       & \textbf{BLEU} & \textbf{ROUGE-L} & \textbf{METEOR} & \textbf{BERTScore} & \textbf{Distinct-1} & \textbf{Distinct-2} & \textbf{Ensemble} \\
    \midrule
    \multirow{16}[8]{*}{SG} & \multirow{4}[2]{*}{M-BERT} & fr->en & 2.62  & 18.8  & 0.102 & 0.893 & 0.907 & 0.986 & 31.131 \\
          &       & fr->de & 2.47  & 14.1  & 0.134 & 0.721 & 0.948 & 0.995 & 34.44 \\
          &       & fr->es & 1.66  & 12.6  & 0.11  & 0.72  & 0.944 & 0.994 & 30.191 \\
          &       & fr->zh & 4.83  & 24.6  & -     & 0.659 & 0.996 & 0.42  & 26.564 \\
\cmidrule{2-10}          & \multirow{4}[2]{*}{XLM} & fr->en & 3.19  & 19.6  & 0.102 & 0.895 & 0.961 & 0.993 & 32.028 \\
          &       & fr->de & 3.21  & 25.7  & 0.141 & 0.729 & 0.958 & 0.99  & 29.395 \\
\bottomrule
    \end{tabular}%
    }
  \caption{The whole results under the French centric cross-lingual evaluation scenarios.}
  \label{tab:cross fr all}%
\end{table*}%

\begin{table*}[htbp]
  \renewcommand\arraystretch{0.6}
  \centering
  \footnotesize
\resizebox{0.93\textwidth}{!}{
    \begin{tabular}{cclccccccc}
    \toprule
    \multirow{2}[4]{*}{\textbf{Task}} & \multirow{2}[4]{*}{\textbf{Model}} & \multicolumn{1}{c}{\multirow{2}[4]{*}{\textbf{Language}}} & \multicolumn{3}{c}{\textbf{N-gram-based}} & \textbf{Embedding-based} & \multicolumn{2}{c}{\textbf{Diversity}} & \textbf{Ours} \\
\cmidrule{4-10}          &       &       & \textbf{BLEU} & \textbf{ROUGE-L} & \textbf{METEOR} & \textbf{BERTScore} & \textbf{Distinct-1} & \textbf{Distinct-2} & \textbf{Ensemble} \\
    \midrule
    \multirow{16}[8]{*}{SG} & \multirow{4}[2]{*}{M-BERT} & es->en & 2.12  & 18.4  & 0.1   & 0.892 & 0.909 & 0.985 & 31.191 \\
          &       & es->de & 2.21  & 13.7  & 0.134 & 0.718 & 0.93  & 0.993 & 34.32 \\
          &       & es->fr & 1.42  & 13.4  & 0.194 & 0.72  & 0.931 & 0.993 & 31.622 \\
          &       & es->zh & 4.51  & 24.2  & -     & 0.658 & 0.995 & 0.337 & 26.406 \\
\cmidrule{2-10}          & \multirow{4}[2]{*}{XLM} & es->en & 3.12  & 19.6  & 0.103 & 0.895 & 0.966 & 0.996 & 31.257 \\
          &       & es->de & 2.83  & 25    & 0.139 & 0.727 & 0.962 & 0.994 & 30.004 \\
\bottomrule
    \end{tabular}%
    }
  \caption{The whole results under the Spanish centric cross-lingual evaluation scenarios.}
  \label{tab:cross es all}%
\end{table*}%

\begin{table*}[htbp]
  \renewcommand\arraystretch{0.6}
  \centering
  \footnotesize
\resizebox{0.93\textwidth}{!}{
    \begin{tabular}{cclccccccc}
    \toprule
    \multirow{2}[4]{*}{\textbf{Task}} & \multirow{2}[4]{*}{\textbf{Model}} & \multicolumn{1}{c}{\multirow{2}[4]{*}{\textbf{Language}}} & \multicolumn{3}{c}{\textbf{N-gram-based}} & \textbf{Embedding-based} & \multicolumn{2}{c}{\textbf{Diversity}} & \textbf{Ours} \\
\cmidrule{4-10}          &       &       & \textbf{BLEU} & \textbf{ROUGE-L} & \textbf{METEOR} & \textbf{BERTScore} & \textbf{Distinct-1} & \textbf{Distinct-2} & \textbf{Ensemble} \\
    \midrule
    \multirow{16}[8]{*}{SG} & \multirow{4}[2]{*}{M-BERT} & zh->en & 2.33  & 18.1  & 0.096 & 0.895 & 0.945 & 0.993 & 31.242 \\
          &       & zh->de & 2.08  & 13    & 0.125 & 0.716 & 0.962 & 0.995 & 34.057 \\
          &       & zh->fr & 1.37  & 13.1  & 0.189 & 0.716 & 0.94  & 0.997 & 30.47 \\
          &       & zh->es & 1.5   & 12.5  & 0.111 & 0.716 & 0.957 & 0.997 & 29.987 \\
\cmidrule{2-10}          & \multirow{4}[2]{*}{XLM} & zh->en & 0.77  & 12.3  & 0.074 & 0.882 & 1     & 1     & 33.839 \\
          &       & zh->de & 1.31  & 17.6  & 0.106 & 0.704 & 0.917 & 1     & 30.245 \\
\bottomrule
    \end{tabular}%
    }
  \caption{The whole results under the Chinese centric cross-lingual evaluation scenarios.}
  \label{tab:cross zh all}%
\end{table*}%

\begin{table*}[htbp]
  \renewcommand\arraystretch{0.6}
  \centering
  \footnotesize
\resizebox{0.93\textwidth}{!}{
    \begin{tabular}{cclccccccc}
    \toprule
    \multirow{2}[4]{*}{\textbf{Task}} & \multirow{2}[4]{*}{\textbf{Model}} & \multicolumn{1}{c}{\multirow{2}[4]{*}{\textbf{Language}}} & \multicolumn{3}{c}{\textbf{N-gram-based}} & \textbf{Embedding-based} & \multicolumn{2}{c}{\textbf{Diversity}} & \textbf{Ours} \\
\cmidrule{4-10}          &       &       & \textbf{BLEU} & \textbf{ROUGE-L} & \textbf{METEOR} & \textbf{BERTScore} & \textbf{Distinct-1} & \textbf{Distinct-2} & \textbf{Ensemble} \\
    \midrule
    \multirow{16}[8]{*}{SG} & \multirow{4}[2]{*}{M-BERT} & en->de & 0.06  & 3.2   & 0.053 & 0.694 & 0.92  & 0.991 & 34.856 \\
          &       & en->fr & 0.04  & 3.9   & 0.041 & 0.705 & 0.921 & 0.991 & 32.033 \\
          &       & en->es & 0.06  & 3.5   & 0.04  & 0.707 & 0.918 & 0.991 & 31.123 \\
          &       & en->zh & 0     & 0.2   & -     & 0.542 & 0.919 & 0.991 & 20.574 \\
\cmidrule{2-10}          & \multirow{4}[2]{*}{XLM} & en->de & 0.02  & 7.2   & 0.05  & 0.634 & 0.469 & 0.5   & 18.939 \\
          &       & en->fr & 0.02  & 5.9   & 0.037 & 0.626 & 0.38  & 0.412 & 14.273 \\
\bottomrule
    \end{tabular}%
    }
  \caption{The whole results under the zero-shot evaluation scenarios.}
  \label{tab:zero all}%
\end{table*}%


\label{sec:appendix}


\end{document}